\documentclass[10pt,twocolumn,letterpaper]{article}

\usepackage{cvpr}              %

\usepackage[accsupp]{axessibility}  %
\usepackage{color}
\usepackage{multirow}
\usepackage{bbding}
\usepackage{graphicx}
\usepackage{amsmath}
\usepackage{amssymb}
\usepackage{booktabs}

\usepackage{float} %
\usepackage{overpic}
\definecolor{myGreen}{rgb}{0.78,0.92,0.73}
\definecolor{yellow}{rgb}{0.99,0.99,0.70}
\definecolor{white}{rgb}{1.0,1.0,1.0}
\definecolor{black}{rgb}{0.00,0.00,0.00}

\usepackage{colortbl} %
\usepackage{nicematrix} %
\usepackage{fnpara} %

\newcommand{\para}[1]{\vspace{.05in}\noindent\textbf{#1}}

\definecolor{cvprblue}{rgb}{0.21,0.49,0.74}
\usepackage[pagebackref,breaklinks,colorlinks,allcolors=cvprblue]{hyperref}

\def\paperID{9861} %
\def\confName{CVPR}
\def\confYear{2025}

\title{Fancy123: One Image to High-Quality 3D Mesh Generation via \\Plug-and-Play Deformation}

\author{Qiao Yu$^{1}$~\quad
Xianzhi Li$^{1,2}$~\quad
Yuan Tang$^{1}$~\quad
Xu Han$^{1}$~\quad
Long Hu$^{1\dag}$~\quad
Yixue Hao$^{1}$~\quad
Min Chen$^{3,4}$ \\
$^1$Huazhong University of Science and Technology~\quad
$^2$Guangdong Intelligent Robotics Institute\\
$^3$South China University of Technology~\quad
$^4$Pazhou Laboratory~\quad $^{\dag}$Corresponding author\\
{\tt\small \{qiaoyu\_epic, xzli, yuan\_tang, xhanxu, hulong, yixuehao\}@hust.edu.cn  \quad minchen@ieee.org}
}

\begin{document}
\twocolumn[{
\maketitle

    \vspace{-2em}
\begin{figure}[H]

    \hsize=\textwidth %
    \centering
    \includegraphics[width=1.0\textwidth]{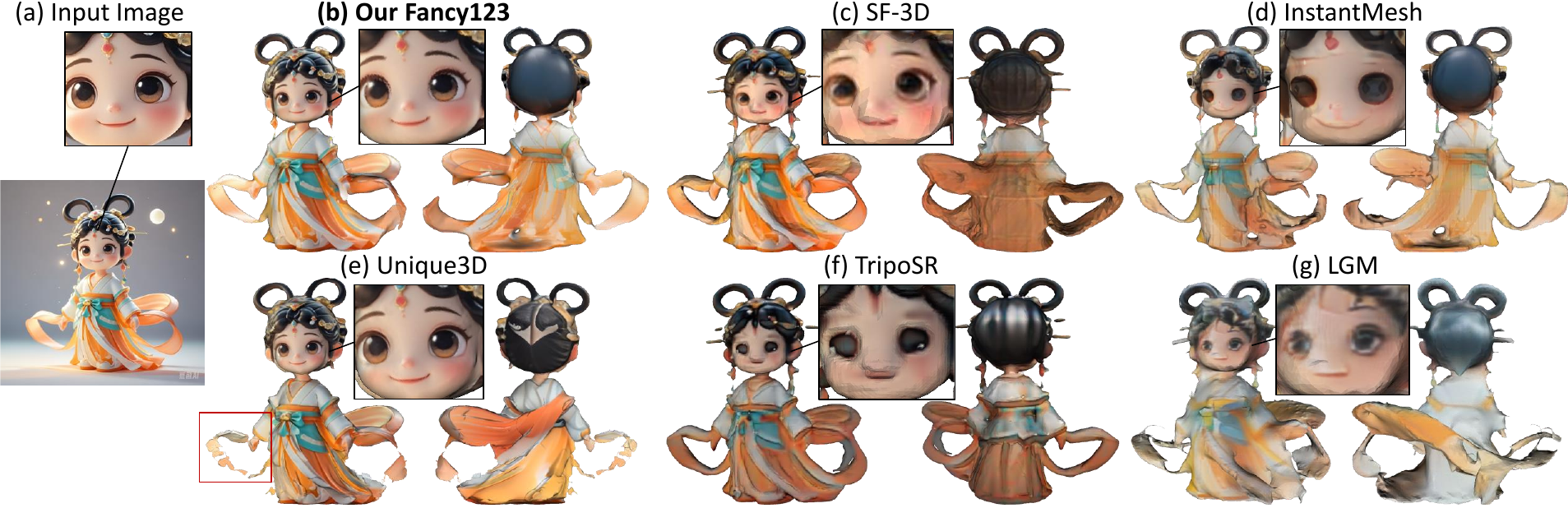}
      \caption{Given a single image, our Fancy123 generates a high-quality 3D mesh in terms of visual appeal, color clarity, and input fidelity.}
  
    \label{fig:teaser}
\end{figure}
}]

\begin{abstract}

Generating 3D meshes from a single image is an important but ill-posed task. Existing methods mainly adopt 2D multiview diffusion models to generate intermediate multiview images, and use the Large Reconstruction Model (LRM) to create the final meshes. However, the multiview images exhibit local inconsistencies, and the meshes often lack fidelity to the input image or look blurry. We propose \textbf{Fancy123}, featuring two enhancement modules and an unprojection operation to address the above three issues, respectively. The appearance enhancement module deforms the 2D multiview images to realign misaligned pixels for better multiview consistency. The fidelity enhancement module deforms the 3D mesh to match the input image. The unprojection of the input image and deformed multiview images onto LRM's generated mesh ensures high clarity, discarding LRM's predicted blurry-looking mesh colors. 
Extensive qualitative and quantitative experiments verify Fancy123's \textbf{SoTA} performance with significant improvement. 
Also, the two enhancement modules are plug-and-play and work at inference time, allowing seamless integration into various existing single-image-to-3D methods. 
Project page: https://github.com/YuQiao0303/Fancy123.

\end{abstract}

\section{Introduction}
\label{sec:intro}

Generating 3D mesh models with a faithful color appearance from a single RGB image (one-image-to-3D, or ``123'') is crucial for applications such as metaverse~\cite{metaverse}, digital twin~\cite{digital_twin}, and virtual reality, as it reduces the expenses of manual 3D modeling. 
However, a single image provides only a limited perspective of the entire object, making it a non-trivial task to infer the 3D shape in a way that is both visually appealing and physically accurate.

In recent years, the ``123" task has witnessed significant progress.
The milestone work DreamFusion~\cite{poole2022dreamfusion} proposed to leverage the strong prior of 2D diffusion models. To improve plausibility in novel views, multiview diffusion models~\cite{shi2023MVDream,wang2023imagedream} are developed to generate multiview images of an object, which are then used to generate 3D geometry and associated color by NeuS~\cite{wang2021neus,liu2023syncdreamer}, normal fusion~\cite{long2023wonder3d}, Large Reconstruction Model (LRM), etc. 
LRM~\cite{hong2023lrm,tang2024lgm,wang2025crm} is a feed-forward network trained with extensive 3D data to create 3D content from a single input or diffusion-generated multiview images within seconds. 

However, existing methods suffer from the following issues: (1) \textbf{Inconsistency among intermediate multiview images}: pixel positions in local regions across different views often do not align. (2) \textbf{Low fidelity to the input}: the generated mesh does not closely resemble the input image; see an example in \cref{fig:teaser} (d) with a slimmer mesh than the input image. (3) \textbf{Blurry coloration}: the generated mesh often look blurry; see \cref{fig:teaser} (c,d,f,g). This blurriness is partly due to the above-mentioned multiview inconsistency. Though each individual view's image is clear, their misaligned regions provide ambiguous references for subsequent mesh color prediction, thus resulting in blurring.

\textbf{To address multiview inconsistency}, existing works encourage diffusion models to generate more consistent images via deliberately designed network structures~\cite{shi2023MVDream,liu2023syncdreamer}, training strategies~\cite{tang2024cycle3d}, and sampling strategies~\cite{shi2023zero123plus}.
However, due to the black-box nature of neural networks, such solutions are uncontrollable and indirect, so the inconsistency persists. In contrast, we address multiview inconsistency from a new perspective: directly correcting inconsistent regions in the already-generated multiview images. To do so, we propose to utilize \textbf{2D image deformation} to explicitly align misaligned pixels by slightly repositioning them. The key idea is to optimize grid-based 2D deformation fields~\cite{deform_2D_1,deform_2d_2} for multiview images given an initial mesh generated by LRM. The optimization objective is that, when the deformed images are unprojected onto the mesh and the mesh is rendered, the rendered images closely resemble the deformed images. A good nature of this explicit 2D deformation is that it works in inference time, and thus can be embedded as a plug-and-play module in any multiview-to-3D method for inconsistency correction.

\textbf{The fidelity issue} is relatively less-explored so far. To address this, we propose to explicitly \textbf{deform the 3D mesh} to match the input image. Specifically, we parameterize the mesh deformation by a Jacobian field~\cite{aigerman2022jacobian,yoo2024apap,Gao_2023_TextDeformer}, and we optimize this field to minimize the difference between the mesh's rendering result and the input image.    
Similar to our 2D deformation, this 3D deformation module is also plug-and-play and works in inference time, allowing seamless integration into any single-image-to-3D method. %

\begin{figure*}[t]
    \centering
    \includegraphics[width=1.0\linewidth]{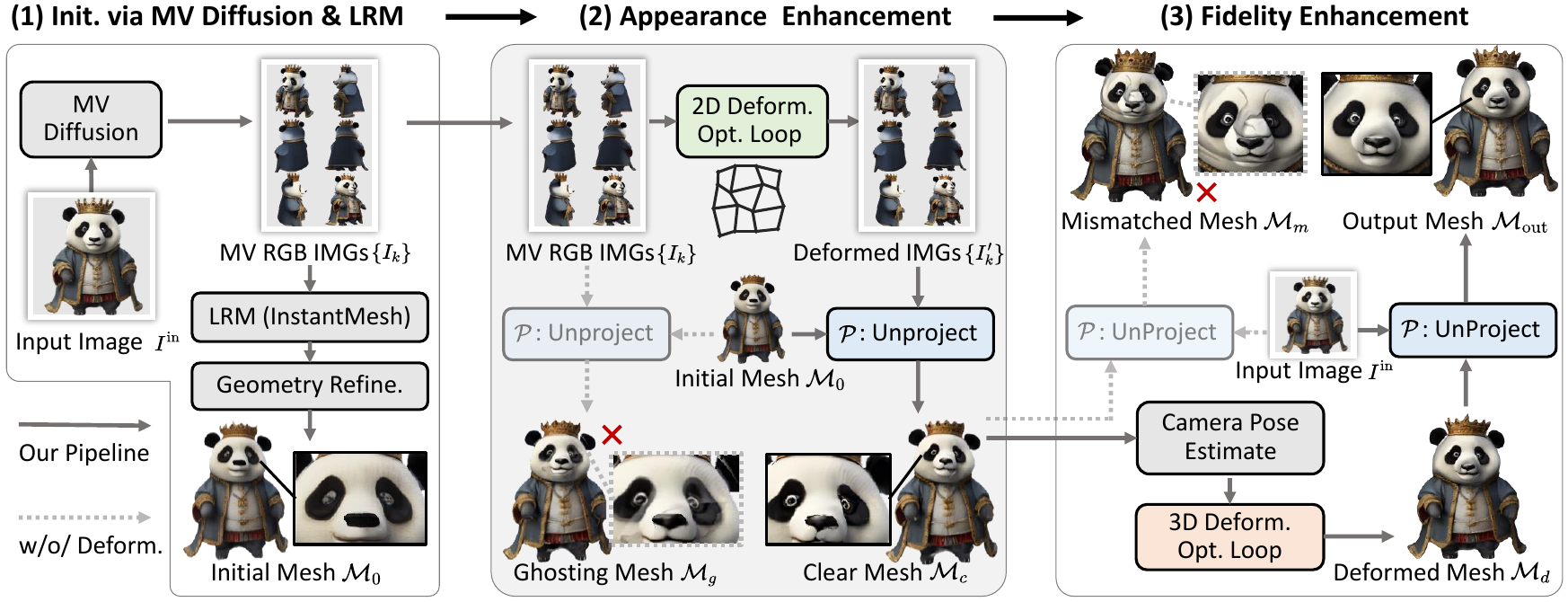}
     \caption{Our pipeline: \textbf{(1)} Initialization via multiview diffusion and LRM: given an input image $I^\text{in}$, use a multiview (MV) diffusion model to generate MV images $\{I_k\}$, use an LRM to create a mesh conditioned on $\{I_k\}$, and then refine mesh geometry to obtain the initial mesh $\mathcal{M}_0$. \textbf{(2)} Appearance enhancement: optimize 2D deformation fields to deform $\{I_k\}$ to $\{I'_k\}$ to ensure MV consistency when unprojecting $\{I'_k\}$ onto $\mathcal{M}_0$ to get $\mathcal{M}_c$, avoiding ghosting as in $\mathcal{M}_g$. \textbf{(3)} Fidelity Enhancement: estimate camera parameters of $I^\text{in}$, based on which deform $\mathcal{M}_c$ into $\mathcal{M}_d$ to match $I^\text{in}$, and finally unproject $I^\text{in}$ onto the $\mathcal{M}_d$ to obtain the final mesh $\mathcal{M}_\text{out}$, avoiding mismatching as in $\mathcal{M}_m$.
    }
    \label{fig:method}
\end{figure*}

To address \textbf{blurry coloration}, a recent work Unique3D~\cite{wu2024unique3d} archives impressive clarity (see \cref{fig:teaser} (e)), largely attributed to its novel designs. However, we find that the often-overlooked operation of ``\textbf{unprojecting multiview images onto the mesh}'' plays a surprisingly significant role in enhancing the clarity, which may have been underestimated in the original work. Specifically, each mesh vertex's color is calculated by the weighted sum of corresponding pixel colors in multiview images. Considering other methods use non-blurry multiview images but still produce blurry meshes, we assume a key point is that methods like LRM do not effectively map colors from images to mesh. On the contrary, unprojection can almost losslessly perform this mapping. %
Luckily, the previous 2D and 3D deformation paves the path for the unprojection of the multiview images and the input image, respectively. Without 2D deformation, directly unprojecting the multiview images to the mesh would cause a ghosting effect due to inconsistency; see ``Ghosting Mesh $\mathcal{M}_g$" in~\cref{fig:method} (2). Similarly, without 3D deformation, directly unprojecting the input image to the mesh would cause artifacts, since the mesh does not match the image; see ``Mismatched Mesh $\mathcal{M}_m$" in~\cref{fig:method} (3).

Building on the above, \textbf{we formulate a new pipeline named Fancy123} with three key steps: (1) given a single image, utilize a multiview diffusion model to generate multiview images, and then quickly reconstruct an initial mesh via an LRM; (2) an appearance enhancement module, which deforms the 2D multiview images to correct inconsistency and unprojects them onto the mesh; and (3) a fidelity enhancement module, which deforms the 3D mesh to match the input image, and unprojects the input onto the mesh.

In summary, our contribution lies in proposing: (1) a new pipeline namely \textit{Fancy123} for enhancing ``123" generation quality, featuring two plug-and-play modules and highlighting the operation of unprojection to avoid blurriness; (2) an appearance enhancement module to correct multiview inconsistency via 2D image deformation; and (3) a fidelity enhancement module to improve the mesh's similarity to the input image via 3D mesh deformation. 
Qualitative and quantitative experiments demonstrate that our Fancy123's \textbf{SoTA} performance by significantly outperforming baseline methods. The effects of each module are verified in our ablation and backbone-replacement experiments. In particular, the two plug-and-play enhancement modules can be seamlessly integrated into various existing ``123" methods.

\section{Related work}
\label{sec:rw}

\para{Approaches for one-image-to-3D}. \
Benefiting from diffusion models~\cite{DDPM} and large-scale 3D datasets~\cite{objaverseXL}, the task of reconstructing colored 3D mesh from a single image has made significant progress~\cite{repaint123,envision3d}.
The pioneer work DreamFusion~\cite{poole2022dreamfusion} utilizes 2D diffusion models for 3D generation via score distillation sampling. To improve novel-view plausibility, multiview~\cite{shi2023MVDream,wang2023imagedream} or view-conditioned~\cite{liu2023zero1to3} diffusion models are developed. With their generated multiview images as reference, 3D meshes can be generated via NeuS~\cite{wang2021neus,liu2023syncdreamer}, normal fusion~\cite{long2023wonder3d}, etc. 
To improve speed, researchers propose LRM~\cite{hong2023lrm}, which is a feed-forward network trained with extensive 3D data for 3D generation, given single image or diffusion-generated multiview images as input. 
Besides, 3D diffusion models~\cite{lan2024ln3diff} or video diffusion models~\cite{yang2024hi3d} are also used for 3D generation, yet not as popular as diffusion and LRM so far. 
While most existing methods yield 
blurry-looking meshes, Unique3D~\cite{wu2024unique3d} achieves a clear appearance by unprojecting multiview images to the mesh, although its geometry reconstruction is slower and often yields artifacts compared to LRM. 

\para{Approaches for multiview consistency.} \
To improve consistency among generated multiview images, existing methods mainly focus on network structures, training strategies, and sampling strategies of multiview diffusion models. MVDream~\cite{shi2023MVDream} proposed a cross-view attention mechanism, and SyncDreamer~\cite{liu2023syncdreamer} designed a 3D-aware feature attention mechanism by unprojecting 2D features to a 3D volume. Zero123++~\cite{shi2023zero123plus} argues for replacing Stable-Diffusion's default scaled-linear schedule for sampling with a linear schedule to emphasize global consistency over local details. Recently, ConsistNet~\cite{yang2023consistnet} proposes a trainable plug-in block to enhance multiview consistency. Cycle3D~\cite{tang2024cycle3d} proposes to iteratively repeat 2D diffusion sampling and 3D feed-forward generation network. 
However, multiview inconsistency caused by pixel misalignment still exists. %

\para{Approaches for 3D mesh deformation.} \
3D Mesh deformation is a fundamental topic in computer graphics, widely applied in 3D generation~\cite{wang2018pixel2mesh,cmrKanazawa18}, interactive shape modeling~\cite{igarashi2005ARAP}, animation, etc. 
Mesh deformation was once commonly used in 3D generation, where researchers deformed an initial mesh (e.g. an ellipsoid) to their desired shape~\cite{ucmrGoel20,umr2020,SMR}. With the emergence of implicit fields~\cite{onet,imnet,deepsdf}, NeRF~\cite{mildenhall2020nerf}, and 3DGS~\cite{3DGS}, this practice has become less common. However, mesh deformation remains crucial in shape editing, which focuses on altering the object's shape reasonably, preserving the local shape and semantics, and ensuring global consistency, smoothness, and plausibility. Among such methods~\cite{lipman2005linear,igarashi2005ARAP, sorkine2004laplacian,jakab2021keypointdeformer,kim2023optctrlpoints}, Jacobian-field-based deformation~\cite{aigerman2022jacobian,Gao_2023_TextDeformer,yoo2024apap} has a nature to preserve global plausibility and smoothness.

\section{Method}

\subsection{Overview}
We present our three-step pipeline in \cref{fig:method}.

\noindent (1) \textbf{Initialization via multiview diffusion and LRM}: We adopt an multiview diffusion model to generate high-quality multiview images $\{I_k\}$, based on which an LRM quickly obtains a corresponding initial mesh $\mathcal{M}_0$.
We also follow~\cite{wu2024unique3d,long2023wonder3d} to further refine the geometry by approximating diffusion-generated multiview normal maps.

\noindent (2) \textbf{Appearance enhancement}: Since the diffusion-generated multiview images $\{I_k\}$ typically have a better visual effect than LRM-generated mesh $\mathcal{M}_0$, we propose to unproject $\{I_k\}$ to $\mathcal{M}_0$ to enhance appearance quality. 
Direct unprojection can cause ghosting due to multiview inconsistency; see $\mathcal{M}_g$ in \cref{fig:method} (2). So we deform $\{I_k\}$ to $\{I'_k\}$, unproject $\{I'_k\}$ onto $\mathcal{M}_0$ to get $\mathcal{M}_c$, and optimize deformation parameters to minimize the difference between $\{I'_k\}$ and rendered images of $\mathcal{M}_c$. 

\noindent (3) \textbf{Fidelity enhancement}: With previous steps, the mesh $\mathcal{M}_c$ achieves enhanced appearance, but may lack fidelity to the input image $I^\text{in}$. Direct unprojecting $I^\text{in}$ to $\mathcal{M}_c$ would cause artifacts due to mismatchment between them, see $\mathcal{M}_m$ in \cref{fig:method} (3). To address this, we propose to deform the 3D mesh $\mathcal{M}_c$ to $\mathcal{M}_d$, to minimize the difference between $I^\text{in}$ and the rendered result of $\mathcal{M}_d$, using camera parameters obtained from estimating $I^\text{in}$'s camera pose. This enables us to unproject the input image to the mesh for higher fidelity. %

Our pipeline takes about 62 seconds on an NVIDIA A100 GPU when adopting InstantMesh as the backbone, with multiview diffusion 6 s, LRM 4 s, geometry refinement 27 s in step 1, 10 s in step 2, and 15 s in step 3, respectively. The peak GPU memory usage is about 21M. %

\subsection{Preliminaries}
\label{sec:preliminaries}
Below, we introduce the key operations in our pipeline. When we refer to these operations in later sections, we directly use the symbols defined here, and treat each operation as a whole module without delving into the specific implementation details.

\para{2D deformation.} We deform a 2D image $I$ by a simple grid-deformation-field strategy~\cite{deform_2D_1,deform_2d_2}, denoted as $I' = \mathcal{D}_\text{2D}(I,F)$, where $I'$ and $F$ denote the deformed images and deformation field, respectively. Specifically, we first evenly divide $I$ with a $G\times G$ grid. Then, we define the deformation field $F\in\mathbb{R}^{G\times G\times 2}$ as the offset of the $G\times G$ grid vertices. $I'$ is obtained by moving the pixels in $I$, where each pixel's position offset is calculated by linear interpolation considering its nearby deformation field grid vertex in $F$.

\para{3D deformation.}
We follow \cite{aigerman2022jacobian,yoo2024apap} to adopt Jacobian-field-based 3D mesh deformation, since it naturally facilitates global smoothness and plausibility~\cite{Gao_2023_TextDeformer}. Given a mesh $\mathcal{M} = (\mathbf{V},\mathbf{T})$ with $V$ vertices $\mathbf{V}\in \mathbb{R}^{V \times 3}$ and $T$ triangle faces $\mathbf{T} \in \{1,...,V\}^{T\times 3}$, we deform it by a Jacobian field $\mathbf{J}$. The output deformed mesh is denoted as $\mathcal{M}_d = (\mathbf{V}',\mathbf{T})$, where $\mathbf{V}' \in \mathbb{R}^{V \times 3}$ are the re-positioned vertices and the faces remain the same. We denote this 3D deformation process as $\mathcal{M}_d = \mathcal{D}_\text{3D}(\mathcal{M}, \mathbf{J})$. 
The Jacobian field $\mathbf{J} = \{\mathbf{J}_{t}|t \in \mathbf{T}\}$ is a learnable representation for the mesh, defined as a set of per-face Jacobians $\mathbf{J}_{t} \in \mathbb{R}^{3 \times 3} $. $\mathbf{J}_t$ is usually initialized as:%
\begin{equation}
    \mathbf{J}_{0, t}=\boldsymbol\nabla_t \mathbf{V},
    \label{equ:J0}
\end{equation}
where $\boldsymbol\nabla_t $ denotes the gradient operator of triangle $t$.
Given $\mathbf{J}$, the coordinates of repositioned vertices $\mathbf{V}'$ can be calculated by solving the following Poisson's equation:
\begin{equation}
    \mathbf{V}'=\underset{\mathbf{V}}{\text{arg}\operatorname*{min}}\|\mathbf{LV}-\boldsymbol{\nabla}^T\mathcal{A}\mathbf{J}\|^2,
\end{equation}
where $\mathbf{L}\in\mathbb{R}^{V\times V}$ is the Laplacian operator,  $\boldsymbol{\nabla}$ is a stack of per-face gradient operators, and $\mathcal{A} \in \mathbb{R}^{3T \times 3T}$ is the mass matrix. 
For more details, please refer to APAP~\cite{yoo2024apap}.

\para{Unprojection.} 
We denote the process of unprojecting multiview images $\{I_k\}$ to a mesh $\mathcal{M}$ as $\mathcal{M}' = \mathcal{P}(\mathcal{M}, \{I_k\},\{\pi_k\})$, where $\mathcal{M}'$ denotes the output mesh, and $\pi_k$ denotes the associated camera parameters of image $I_k$. Specifically, we follow Unique3D~\cite{wu2024unique3d} to calculate each mesh vertex's color as the weighted sum of all corresponding pixels in $\{I_{k}\}$. The weights are the cosine similarity of the 3D vertex's normal direction and the view $k$'s camera direction. See Unique3D~\cite{wu2024unique3d} for more details.

\para{Differentiable rendering.} We denote the differentiable rendering process as $I^R = \mathcal{R}(\mathcal{M},\pi)$, where $\mathcal{M}$ is the input colored mesh, $\pi$ is the camera parameters, and $I^R$ is the rendered image.

\subsection{Initialization via multiview diffusion and LRM}
\label{sec:init}
Given an input image $I^\text{in}$, we generate multiview images $\{I_k\}$ by a multiview diffusion model, and then create a mesh by an LRM conditioned on $\{I_k\}$. Among existing open-source multi-view-input LRM, we adopt the latest InstantMesh~\cite{xu2024instantmesh} with its associated multiview diffusion model fine-tuned from~\cite{shi2023zero123plus}, since it's effective and fast. Other methods can also work. After that, we follow~\cite{wu2024unique3d,long2023wonder3d} to further refine the mesh's geometry, by optimizing mesh vertex coordiantes using multiview normal maps generated via ~\cite{shi2023zero123plus}. The output mesh of this step is denoted as $\mathcal{M}_0$.

\subsection{Appearance enhancement}
The multiview images $\{I_{k}\}$ typically exhibit superior visual quality compared to mesh $\mathcal{M}_{0}$, so we follow Unique3D~\cite{wu2024unique3d} to unproject $\{I_{k}\}$ onto $\mathcal{M}_0$ to enhance appearance. However, a direct unprojection would cause ghosting due to inconsistencies among $\{I_{k}\}$; see $\mathcal{M}_g$ in \cref{fig:method} (b). 

\begin{figure}[t]
    \centering
    \includegraphics[width=1.0\linewidth]{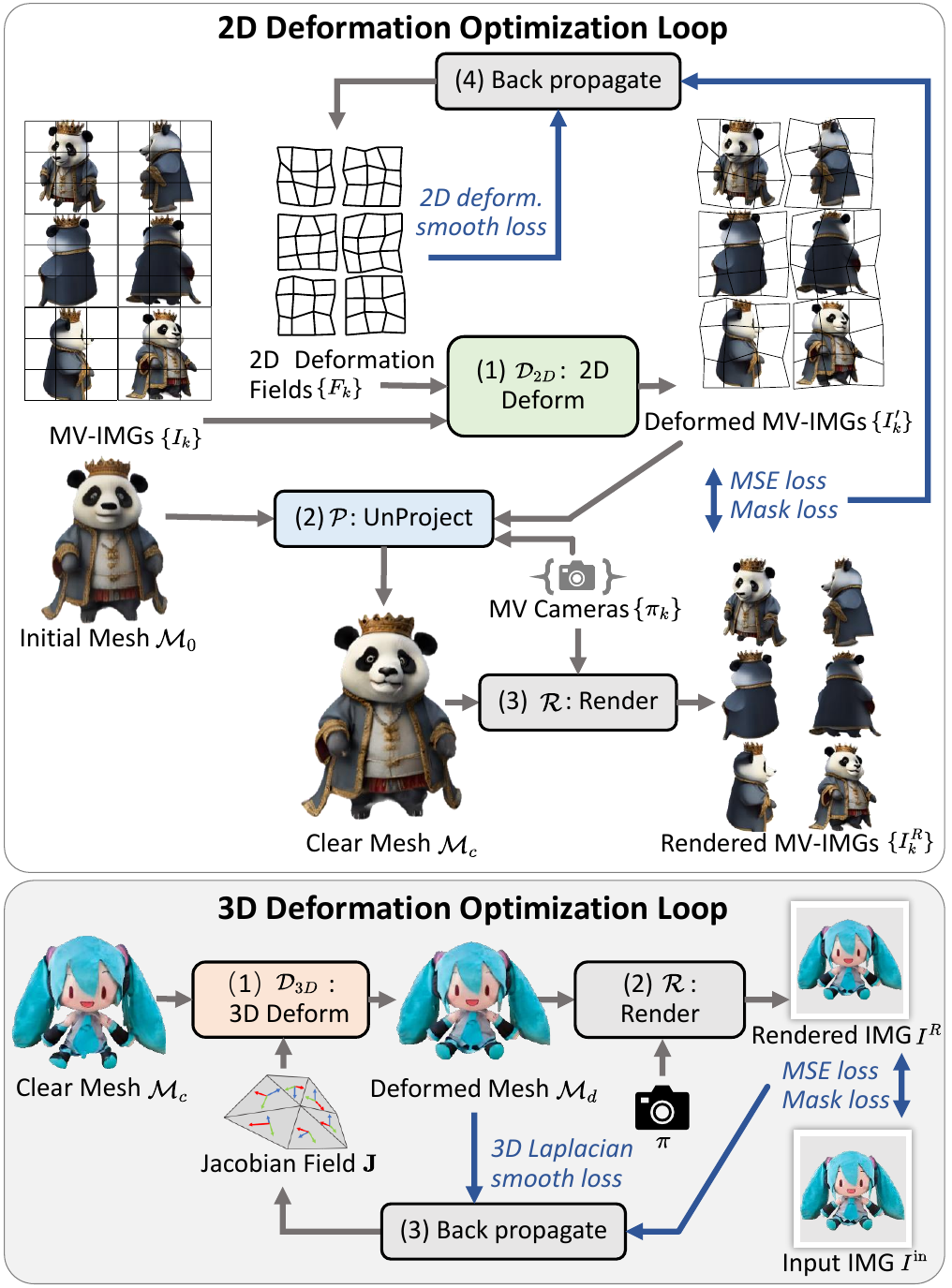}
    \caption{2D Deformation Optimization Loop: optimize 2D deformation fields to deform the MV images to enhance multiview consistency. 3D Deformation Optimization Loop: Optimize a Jacobian field to deform the 3D mesh to resemble the input image.}
    \label{fig:2D_3D_deform_method}
\end{figure}

To address this issue, we propose a 2D deformation optimization loop, as illustrated at the top of \cref{fig:2D_3D_deform_method}. Given an initial mesh $\mathcal{M}_0$ and associated multiview images $\{I_{k}\}$, the idea is to iteratively optimize a 2D deformation field $F_{k}$ for each $I_{k}$, so that when unprojecting the deformed images $\{I'_{k}\}$ onto $\mathcal{M}_0$, the result mesh $\mathcal{M}_{c}$ would closely resemble $\{I'_{k}\}$. Since $\{I'_{k}\}$ are clear without blurring or ghosting, $\mathcal{M}_{c}$ would have a clear appearance too.%

Specifically, we first initialize $\{F_{k}\}$ to zero. 
Then, for each optimization iteration, we:

(1) deform $\{I_{k}\}$ by $\{F_{k}\}$ to obtain deformed multiview images: $\{I'_{k}\} = \{\mathcal{D}_\text{2D}(I_k,F_k)\}$; 

(2) unproject deformed images $\{I'_{k}\}$ onto mesh $\mathcal{M}_0$ with $\{I_{k}\}$'s associated camera parameters $\{\pi_k\}$ to obtain mesh  
$ \mathcal{M}_c = \mathcal{P}(\mathcal{M}_0, \{I'_k\},\{\pi_k\}) $, where $\{\pi_k\}$ is determined by the adopted multiview diffusion model;

(3) differentiably render $\mathcal{M}_c$ by camera parameters ${\pi_k}$: $\{I^R_k\} = \{\mathcal{R}(\mathcal{M}_c, \pi_k)\}$;

(4) calculate the MSE loss, mask loss, and deformation smooth loss for backpropagation to update the deformation field offsets $\{F_k\}$. The MSE loss is defined as:

\begin{equation}
\mathcal{L}_{\text{MSE}_1}=\sum_k\left\|{C}'_k-C_k^{R}\right\|_2^2. 
\end{equation}
${C}'_k$ and ${C}^R_k$ denote the RGB channel values of view $k$'s deformed image $I'_k$ and rendered image $I^R_k$, respectively. 

The mask loss is defined as:
\begin{equation}
\mathcal{L}_{\text{mask}_1}=\sum_k\left\|{M}'_k-M_k^{R}\right\|_2^2.
\end{equation}
${M}'_k$ and ${M}^R_k$ denote the alpha channel values of view $k$'s deformed image $I'_k$ and rendered image $I^R_k$, respectively.

The smooth loss is defined as :

\begin{equation}
\begin{aligned}
&\mathcal{L}_{\text{smooth}_\text{2D}} = \frac{1}{K} \sum_{k=1}^{K} \sum_{i=1}^{G-1} \sum_{j=1}^{G} | F_k(i,j) - F_k(i+1,j) | \\
&\quad + \frac{1}{K} \sum_{k=1}^{K} \sum_{i=1}^{G} \sum_{j=1}^{G-1} | F_k(i,j) - F_k(i,j+1) |.
\end{aligned}
\end{equation}
$K$ is the total number of views, and $G$ is $F_k$'s resolution. This loss tries to make the offset value of each grid vertex $F_k(i,j)$ to be close to its neighbors $F_k(i+1,j)$ and $F_k(i,j+1)$, so as to make the deformation field smooth.

The final loss for the whole appearance enhancement module is the weighted sum of the above losses, where detailed weight values $w_1$-$w_3$ are in supplementary file:
\begin{equation}
    \mathcal{L}_\text{appearence} = w_1  \mathcal{L}_{\text{MSE}_1} + 
 w_2 \mathcal{L}_{\text{mask}_1} + 
 w_3 \mathcal{L}_{\text{smooth}_\text{2D}}.
\end{equation}
After 100 (experimentally determined) iterations, we can effectively alleviate the ghosting or blurry effect due to multiview inconsistency in the output mesh $\mathcal{M}_c$.

\subsection{Fidelity enhancement}
\label{sec:fidelity}
To further improve our generated mesh $\mathcal{M}_c$'s fidelity to the input image $I^\text{in}$, we propose a fidelity enhancement module. The idea is to (1) estimate the camera parameters $\pi$ of $I^\text{in}$ for subsequent optimization; %
(2) deform $\mathcal{M}_c$ to $\mathcal{M}_d$ by optimizing a jacobian field, so that when rendering $\mathcal{M}_d$ from $\pi$, the result $\mathcal{R}(\mathcal{M}_c, \pi)$ would closely resemble $I^\text{in}$; and (3) unproject $I^\text{in}$ onto $\mathcal{M}_d$.

\para{Camera pose estimation.} 
The objective of 3D mesh deformation is to make the deformed mesh closely resemble the input image $I^\text{in}$ when observed from camera parameters $\pi$. Therefore, to reduce the difficulty of deformation, we should choose $\pi$ to ensure that the undeformed mesh $\mathcal{M}_c$ already has good correspondence with $I^\text{in}$.
Some single-image-to-3D methods~\cite{wu2024unique3d} use elevation and azimuth angles relative to $I^\text{in}$, so simply setting these two angles to zero can align $\mathcal{M}_c$ with $I^\text{in}$. However, our adopted IntantMesh~\cite{xu2024instantmesh} uses relative azimuth and absolute elevation angles. When $I^\text{in}$ has a big absolute elevation angle, such default camera setting may be unsuitable, necessitating camera pose estimation to find an appropriate 
$\pi$. Specifically, we perform a coarse-to-fine parameter search over all possible elevation angles and select the $\pi$ that minimizes the LPIPS~\cite{lpips} score: $\text{LPIPS}(I^\text{in},\mathcal{R}(\mathcal{M}_c,\pi))$, where LPIPS is a metric indicating the difference between two images. We also add a small optimization loop to further optimize $\pi$. See our supplementary file for more details.

\para{3D deformation optimization.}
Given the colored mesh $\mathcal{M}_c$, the input image $I^\text{in}$, and the camera parameters $\pi$, we use an optimization loop to deform $\mathcal{M}_c$ to approximate $I^\text{in}$, as shown in the bottom of \cref{fig:2D_3D_deform_method}. First, we follow \cref{equ:J0} to initialize the Jacobian field $\mathbf{J}$. Then, for each iteration, we:

(1) deform $\mathcal{M}_c$ to be $\mathcal{M}_d = \mathcal{D}_\text{3D}(\mathcal{M}_c, \mathbf{J})$, 

(2) render $\mathcal{M}_d$ by $\pi$ to get $I^R = \mathcal{R}(\mathcal{M}_d,\pi)$, and 

(3) calculate the MSE loss and mask loss between $I^R$ and $I^\text{in}$, and 3D Laplacian smooth loss of $\mathcal{M}_d$ for backpropagation to update $\mathbf{J}$. The MSE loss is defined as:

\begin{equation}
    \mathcal{L}_{\text{MSE}_2}=\left\|{C}^{in}-C^{R}\right\|_2^2,
\end{equation}
where ${C}^{in}$ and ${C}^R$ denote the RGB channel values of the input image $I^{in}$ and rendered image $I^R$, respectively.

The mask loss is defined as :
\begin{equation}
    \mathcal{L}_{\text{mask}_2}=\left\|{M}^{in}-M^{R}\right\|_2^2,
\end{equation}
where ${M}^{in}$ and ${M}^R$ denote the alpha channel values of the input image $I^{in}$ and rendered image $I^R$, respectively.

The 3D Laplacian smooth loss is defined as:
\begin{equation}
\label{equ:3D_lap}
    \mathcal{L}_\text{Lap} = \left\|\mathbf{LV'}\right\|^2,
\end{equation}
where $\mathbf{L}$ and $\mathbf{V}'$ denote the discrete Laplacian operator and the vertices of deformed mesh $\mathcal{M}_d$, respectively.

The final loss of the fidelity enhancement module is the weighted sum of the above losses, where values of $w_4$-$w_6$ are in the supplementary file:
\begin{equation}
    \mathcal{L}_\text{fidelity} = w_4 \mathcal{L}_{\text{MSE}_2} + w_5 \mathcal{L}_{\text{mask}_2}+ w_6\mathcal{L}_\text{Lap}.
\end{equation}

\begin{table}[t]
  \centering
   \resizebox{1.0\linewidth}{!}{
    \begin{NiceTabular}{c|ccccc|cc}
    \toprule
          & \multicolumn{5}{c|}{Appearence Metrics} & \multicolumn{2}{c}{Geometry Metrics} \\
    \midrule
    Method & \multicolumn{1}{c}{FID ↓} & \multicolumn{1}{c}{LPIPS ↓} & \multicolumn{1}{c}{PSNR ↑} & \multicolumn{1}{c}{SSIM ↑} & \multicolumn{1}{c|}{CLIP-Sim. ↑} & \multicolumn{1}{c}{CD ↓} & \multicolumn{1}{c}{F-Score ↑} \\
    \midrule
    
    LGM~\cite{tang2024lgm}  & 86.24 & 0.395 & 13.25 & 0.612 & 0.710 & 0.0223 & 0.8555 \\

    TripoSR~\cite{TripoSR2024} & 68.52 & 0.367 & 13.88 & 0.629 & 0.739 & 0.0164 & 0.8876 \\

    CRM~\cite{wang2025crm}  & 87.74 & 0.388 & 13.66 & 0.620 & 0.729 & 0.0161 & \cellcolor[rgb]{ 1,  .612,  .6} 0.9064 \\
        
    InstantMesh~\cite{xu2024instantmesh} & \cellcolor[rgb]{ 1,  .788,  .78} 46.16 & 0.349 & 13.74 & \cellcolor[rgb]{ 1,  .788,  .78} 0.634 & \cellcolor[rgb]{ 1,  .788,  .78} 0.800 & 0.0153 & \cellcolor[rgb]{ 1,  .788,  .78} 0.9035 \\
    
    Unique3D~\cite{wu2024unique3d}  & 58.51 & 0.389 & 12.74 & 0.592 & 0.764 & 0.0266 & 0.8966 \\
    
    SF3D~\cite{sf3d2024}  & 49.89 & \cellcolor[rgb]{ 1,  .788,  .78} 0.343 & \cellcolor[rgb]{ 1,  .788,  .78} 14.32 & 0.617 & 0.776 & \cellcolor[rgb]{ 1,  .612,  .6} 0.0142 & 0.8090 \\

    \midrule
    Ours  & \cellcolor[rgb]{ 1,  .612,  .6} 37.99 & \cellcolor[rgb]{ 1,  .612,  .6} 0.330 & \cellcolor[rgb]{ 1,  .612,  .6} 14.37 & \cellcolor[rgb]{ 1,  .612,  .6} 0.651 & \cellcolor[rgb]{ 1,  .612,  .6} 0.835 & \cellcolor[rgb]{ 1,  .788,  .78} 0.0151 & \cellcolor[rgb]{ 1,  .612,  .6} 0.9064 \\
    \bottomrule
    \end{NiceTabular}%
    }
   
    \caption{Quantitative comparisons of our method against baseline methods (ranked by initial paper release time) for the single-image-to-3D-mesh task.  }
     \vspace{-1em}
  \label{tab:main_results}%
\end{table}%

\para{Unproject input image.} After the 3D deformation (200 iterations), we can safely unproject $I^\text{in}$ to $\mathcal{M}_d$ to obtain the final output mesh $\mathcal{M}_\text{output} = \mathcal{P}(\mathcal{M}_d, \{I^\text{in}\}, \{\pi\})$.

 \section{Experiments}
\label{sec:exp}

\begin{figure*}[t]
    \centering
    \includegraphics[width=1.0\linewidth]{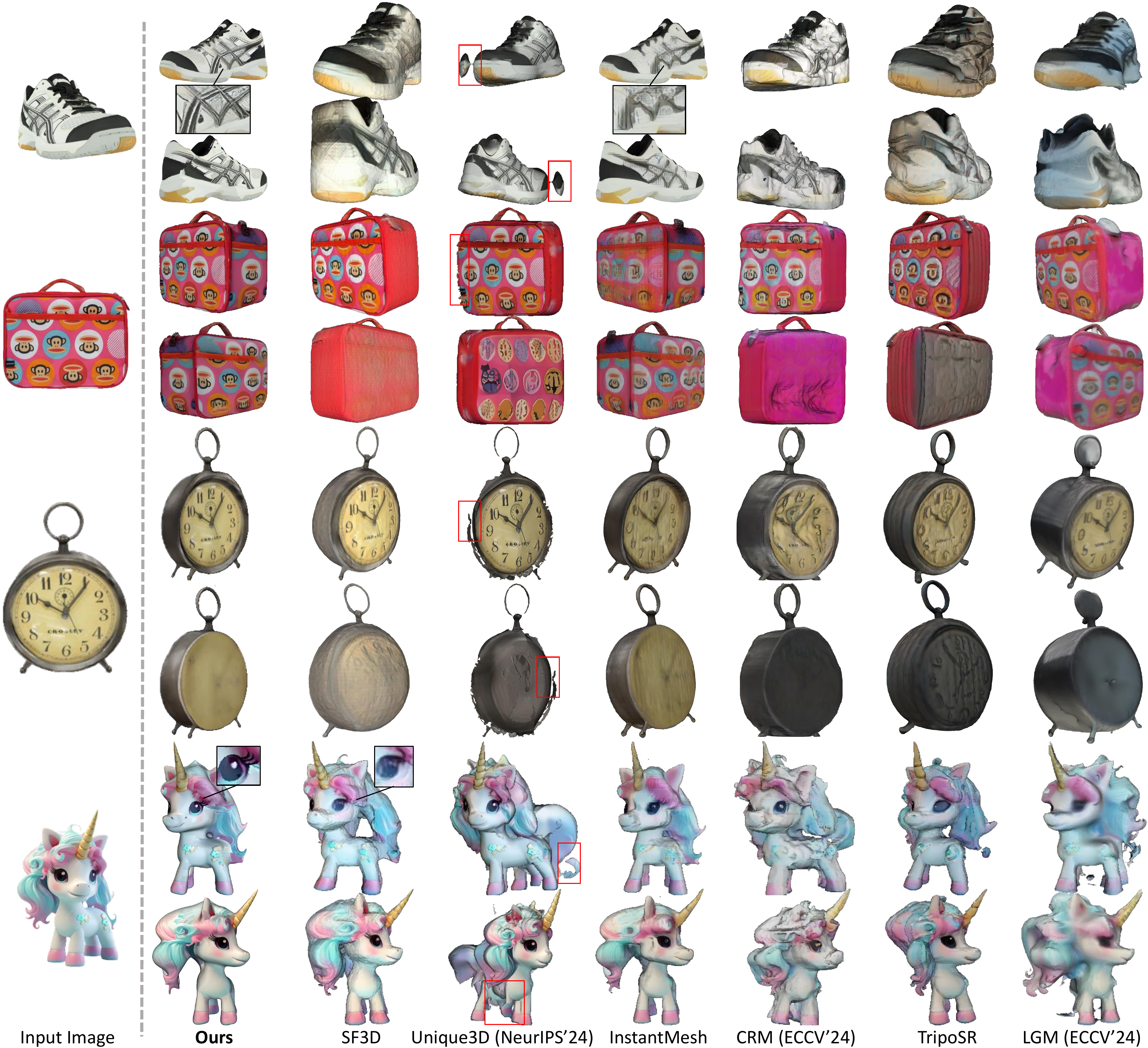}
    \caption{Visual comparisons of our method against baseline methods for the single-image-to-3D-mesh task.}
    \label{fig:main_result}
    \vspace{-1em}
\end{figure*}

\subsection{Comparison with SoTA methods}
\label{sec:compare_sota}
\para{Datasets:} We follow existing works~\cite{xu2024instantmesh,wu2024unique3d} to use the Google Scanned Objects~\cite{downs2022google} (GSO) dataset to evaluate the performance of the ``123'' task. We follow TripoSR~\cite{TripoSR2024} to manually filter the dataset to exclude simple objects like medicine boxes, then randomly select 100 objects, unlike \cite{long2023wonder3d,liu2023syncdreamer} which only use 30. For the selected samples, we render them with a resolution of $1024^2$, an elevation of 0° and an azimuth of -30° as input, considering a non-zero azimuth provides more information about the object's depth. We also provide results of using frontal views (elevation = azimuth = 0°) as input in our supplementary file. Besides, we collect some images from the Internet as input for additional qualitative comparison, e.g. the pony in \cref{fig:main_result}. %

\para{Baselines: } We compare our method against a wide range of SoTA single-image-to-3D methods, including LGM~\cite{tang2024lgm}, TripoSR~\cite{TripoSR2024}, CRM~\cite{wang2025crm},  InstantMesh~\cite{xu2024instantmesh}, Unique3D~\cite{wu2024unique3d}, and SF3D~\cite{sf3d2024}. For LGM, we use the official code to convert the generated 3D Gaussians into meshes.

\para{Evaluation strategy and metrics.} 
We follow existing works~\cite{xu2024instantmesh,wu2024unique3d,sf3d2024} to compare ground truth and generated mesh considering both appearance and geometry. %
Specifically, for appearance comparison, we follow~\cite{wu2024unique3d} to render 24 views for each sample, where the elevation angles are [0,15,30] and the azimuth angles are evenly distributed around the 360 degrees. We compare the rendering results of generated and GT meshes with commonly used metrics for image similarity/distance measuring, including PSNR, SSIM~\cite{SSIM}, LPIPS~\cite{lpips}, FID~\cite{FID}, and Clip-Similarity~\cite{clip}. For 3D geometry evaluation, we follow ~\cite{xu2024instantmesh} to report Chamfer Distance (CD) and F-Score (FS), both reflecting the similarity between GT and generated 3D objects. All meshes are normalized within the size of [-1,1], where 100k points are sampled for each mesh for geometric evaluation.

As mentioned, some methods (InstantMesh and LGM) do not align with the input view by default, so we align them by coarse-to-fine parameter search, and adopt the camera parameters with the lowest LPIPS score for rendering. See our supplementary for more details.

\para{Qualitative results}. 
\cref{fig:main_result} shows the qualitative comparisons. As can be seen, the shoe's texture, the bag's patterns, the clock's numbers, and the pony's eyes contain fine details, making clear coloration particularly challenging. Regarding this, only our method and Unique3D achieved clear and sharp coloration, while other methods exhibited varying degrees of blurriness. However, Unique3D often yields artifacts, as marked by the red boxes, suggesting a lack of stability. 
Furthermore, the input images for the bag and the clock only provide front-view information, making the reconstruction of the back side challenging. Our method produces clear and plausible results. In particular, SF3D and TripoSR show obvious artifacts on the back side, possibly because they do not use intermediate multiview images but directly produce meshes from a single image by LRM.

\para{Quantitative results}. \cref{tab:main_results} lists quantitative results, where our method archives the best performance for 6 out of 7 metrics, particularly all 6 appearance metrics significantly. %

However, we experimentally find that \textbf{comparing GT and generated meshes using these metrics is not reliable}, as it does not align well with human perception.
For example, \cref{fig:ablation} (b-d) shows increasing appearance quality by human perception, but the PSNR scores are 12.76, 12.35, and 12.44, respectively, indicating (b) is the best and (c) is the worst. Other metrics also have the similar issue; see our supplementary file for more examples. 
We analyze that the inconsistency between quantitative metrics and visual judgments is mainly due to: %
(1) All current metrics solely assess fidelity (i.e., the similarity between GT and generated mesh), not plausibility. If two meshes both poorly match the input, the more plausible one deserves a higher score, but existing metrics treat them equally.
(2) Due to the input's lack of perspectives, the GT of this task should be diverse. 
For example, in \cref{fig:main_result}, any color on the back of the red bag is reasonable as long as there are no obvious artifacts. Therefore, evaluating with a single definite GT is inproper.%

Regarding this, LGM~\cite{tang2024lgm} (ECCV'24 Oral) completely discards this evaluation strategy. Many works~\cite{xu2024instantmesh,wu2024unique3d} only use these metrics for comparison with SoTA, but adopt visual comparison for ablation study. \textbf{Here we recommend focusing on the visual comparisons too, and only comparing visual results in our following experiments.}

\subsection{Ablation study}
\begin{figure*}[t]
    \centering
    \includegraphics[width=1.0\linewidth]{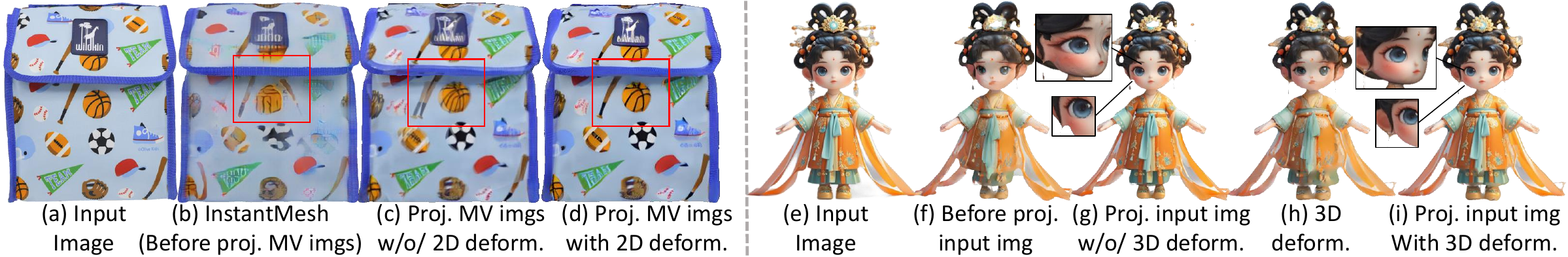}
    \caption{Ablation experiments on our appearance (a-d) and fidelity (e-i) enhancement modules via 2D and 3D deformation, respectively. %
    }
    \label{fig:ablation}
\end{figure*}

\begin{figure}[t]
    \centering
    \includegraphics[width=1.0\linewidth]{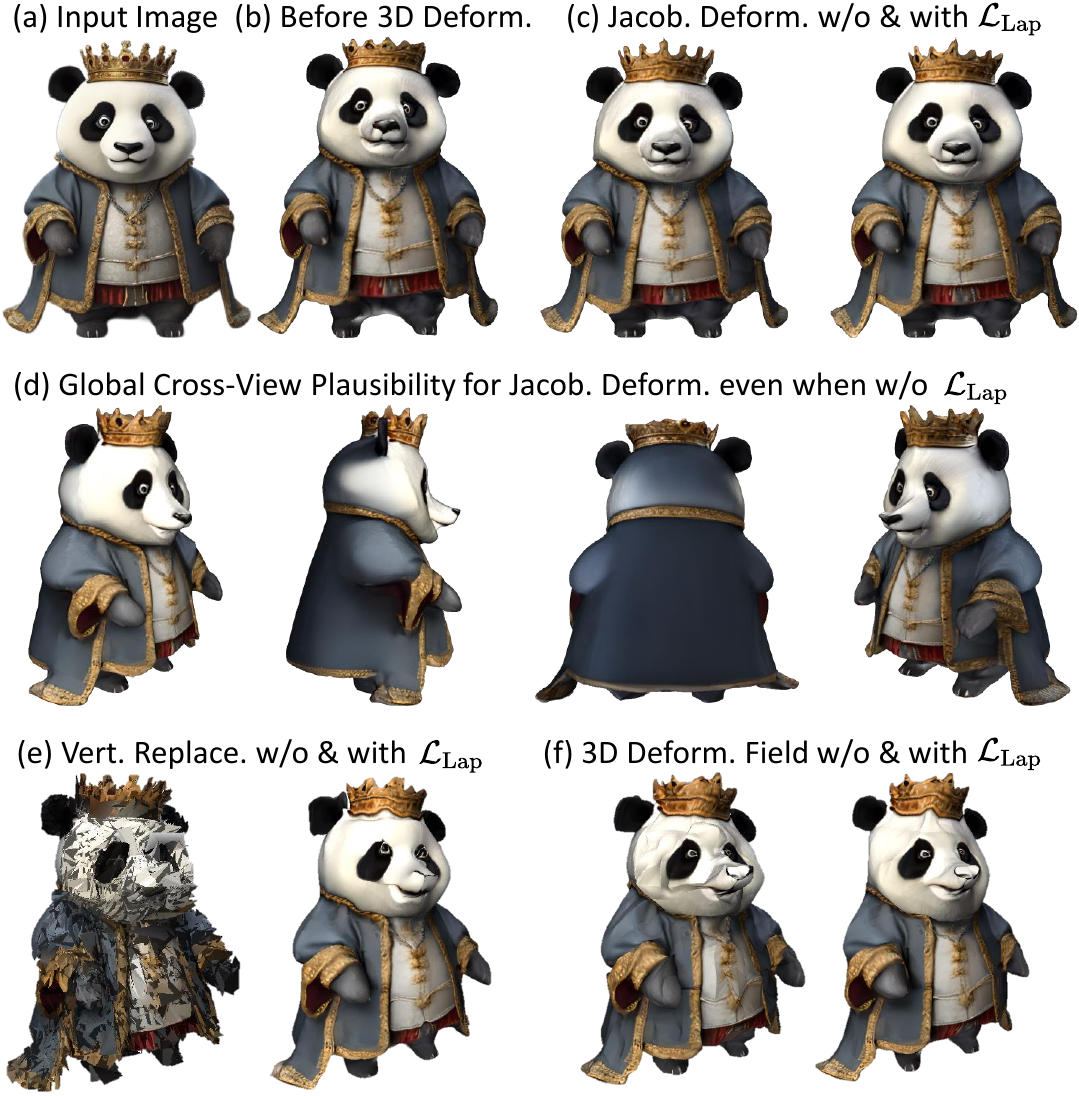}
      \caption{Superiority of Jacobian-based deformation: globally smooth and plausible. When deforming the mesh to resemble the input (a-c), it achieves global plausibility even when w/o Laplacian smooth loss (d), compared to other deformation methods (e-f).}
    \label{fig:3D_deform_ablation}
    \vspace{-1em}
\end{figure}

\begin{figure}[t]
    \centering
    \includegraphics[width=1.0\linewidth]{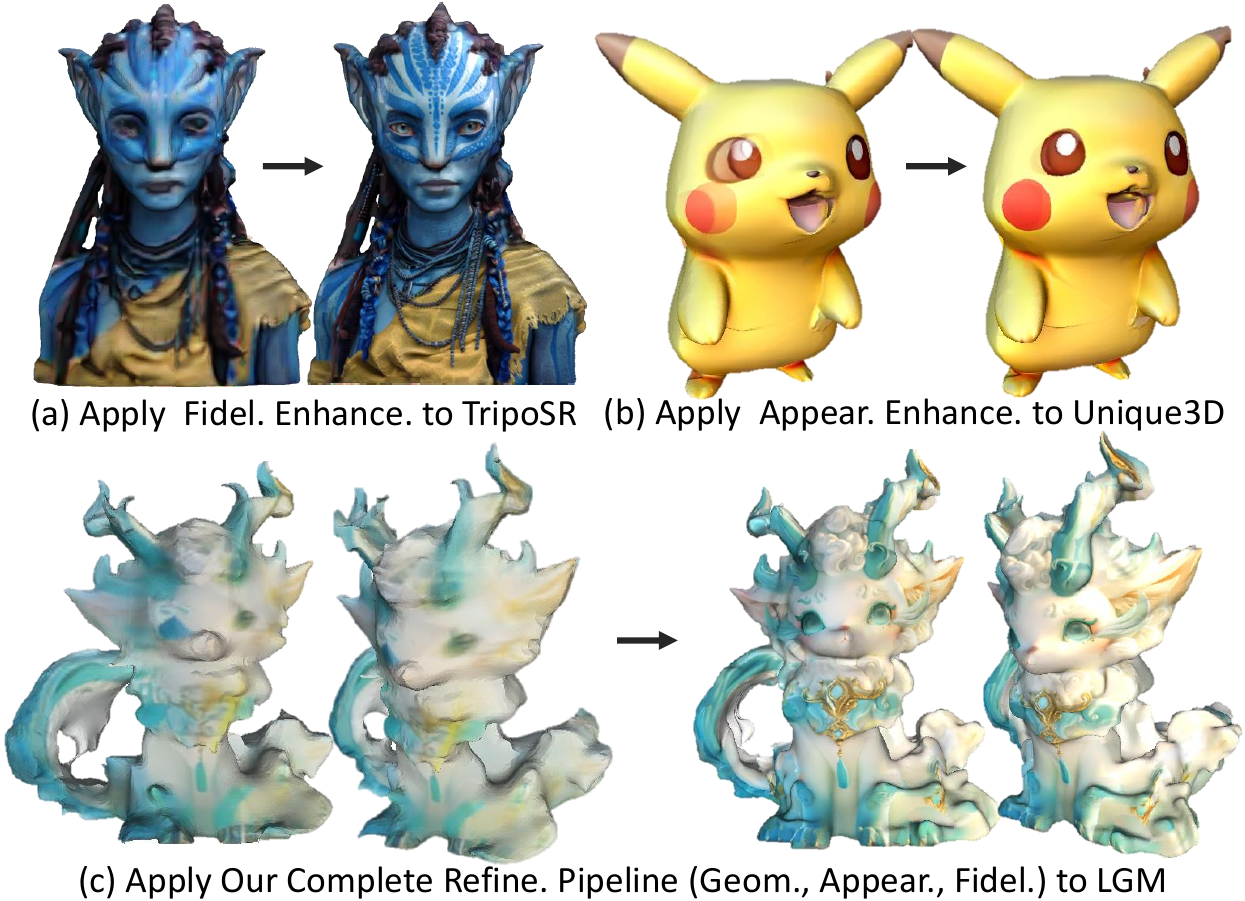}
    
    \caption{
    Improvement is significant when applying our various plug-and-play enhancement modules to other backbones.%
    }
    \label{fig:other_backbone}
    \vspace{-1em}
\end{figure}

\para{Effect of appearance enhancement.}
\cref{fig:ablation} (a)-(d) show the effect of our appearance enhancement by unprojecting deformed multiview images to the mesh:

(b) InstantMesh's original blurry-looking mesh. %

(c) Directly unprojecting multiview images to the meshes may increase clarity, but may cause ghosting due to multiview inconsistency, as marked in red boxes. 

(d) Our 2D deformation effectively mitigated this issue.

Please refer to our supplementary file for more examples.

\para{Effect of fidelity enhancement.}
\cref{fig:ablation} (e)-(i) show the effect of our fidelity enhancement by unprojecting the input image to the deformed mesh:

(f) The meshes after appearance enhancement may lack sufficient similarity to the input image (face too narrow). %

(g) directly unprojecting input image to the mesh causes artifacts (lack of an ear, too wide eye etc.)

(h) Deforming the 3D mesh to approximate the input image improves the fidelity.

(i) Since the mesh in (f) closely matches the input image, we now safely unproject the image to the mesh. This can improve both fidelity and clarity.

\para{Cross-view plausibility of 3D deformation.}
Our 3D deformation only uses the input image for supervision. Will this influence the plausibility of other views? Luckily, our adopted Jacobian-based deformation addresses this issue. 
While making the mesh better resemble the input (\cref{fig:3D_deform_ablation} (a)-(c)), the Jacobian-field-based deformation keeps a high cross-view plausibility (\cref{fig:3D_deform_ablation} (d)) even when without any Laplacian smooth loss, compared to other possible deformation methods such as vertex replacement or 3D deformation field in \cref{fig:3D_deform_ablation} (e)-(f). We provide more details about the two other deformation methods in our supplementary file.%

\para{Backbone replacement.}
Besides InstanceMesh, our proposed appearance and fidelity enhancement modules can be flexibly used in other backbone methods.
In \cref{fig:other_backbone}, significant improvement can be seen when applying (a) our fidelity enhancement to TripoSR~\cite{TripoSR2024}, (b) our appearance enhancement to Unique3D~\cite{wu2024unique3d}, and (c) our complete refinement pipeline to LGM~\cite{tang2024lgm}.

\section{Conclusion}

We propose Fancy123, a novel SoTA framework for enhancing single-image-to-3D-mesh generation quality. It features appearance and fidelity enhancement modules to enhance multiview consistency and fidelity by plug-and-play deformation, and an unprojection operation to ensure clear coloration.
Though promising reconstruction quality has been verified in extensive experiments, Fancy123 still has limitations. 
First, like other methods, its performance is constrained by the adopted multiview diffusion model. 
Second, our 3D deformation is only supervised with RGBA images, thus can sometimes cause artifacts when the object's different semantic parts share similar colors.  
See our supplementary file for an example. %
In the future, we plan to investigate adopting semantic guidance to address this issue, and expand our method to adapt to large-scale scenes. %

\clearpage

  \section*{Acknowledgements}

This work was supported by the China National Natural Science Foundation No. 62202182, Guangdong Basic and Applied Basic Research Foundation 2024A1515010224, and also by the National Key Research and Development Program of China under Grant 2023YFB4503400, the China National Natural Science Foundation No.62176101, 62276109, 62450064, and 62322205.

{
    \small
    \bibliographystyle{ieeenat_fullname}
    \bibliography{main}

\begin{thebibliography}{55}
\providecommand{\natexlab}[1]{#1}
\providecommand{\url}[1]{\texttt{#1}}
\expandafter\ifx\csname urlstyle\endcsname\relax
  \providecommand{\doi}[1]{doi: #1}\else
  \providecommand{\doi}{doi: \begingroup \urlstyle{rm}\Url}\fi

\bibitem[Aigerman et~al.(2022)Aigerman, Gupta, Kim, Chaudhuri, Saito, and Groueix]{aigerman2022jacobian}
Noam Aigerman, Kunal Gupta, Vladimir~G Kim, Siddhartha Chaudhuri, Jun Saito, and Thibault Groueix.
\newblock {Neural Jacobian Fields: learning intrinsic mappings of arbitrary meshes}.
\newblock \emph{ACM Transactions on Graphics (TOG)}, 41\penalty0 (4):\penalty0 1--17, 2022.

\bibitem[AiuniAI(2024)]{unique3d_normal}
AiuniAI.
\newblock Unique3d/app/custom\_models/normal\_prediction.py at main.
\newblock \url{https://github.com/AiuniAI/Unique3D/blob/main/app/custom_models/normal_prediction.py}, 2024.

\bibitem[Boss et~al.(2024)Boss, Huang, Vasishta, and Jampani]{sf3d2024}
Mark Boss, Zixuan Huang, Aaryaman Vasishta, and Varun Jampani.
\newblock {SF3D: Stable Fast 3D Mesh Reconstruction with UV-unwrapping and Illumination Disentanglement}.
\newblock \emph{arXiv preprint arXiv:2408.00653}, 2024.

\bibitem[Chen et~al.(2024)Chen, Zhao, Zheng, Wang, Ling, Tao, and Yin]{deform_2d_2}
Teru Chen, Xingwei Zhao, Guo Zheng, Jiaxin Wang, Qing Ling, Bo Tao, and Zhouping Yin.
\newblock Real-time deformable registration for robotic ultrasound surgical guidance based on vectorized trust-region optimization.
\newblock \emph{IEEE Transactions on Instrumentation and Measurement}, 73:\penalty0 1--11, 2024.

\bibitem[Chen and Zhang(2019)]{imnet}
Zhiqin Chen and Hao Zhang.
\newblock Learning implicit fields for generative shape modeling.
\newblock In \emph{CVPR}, pages 5939--5948, 2019.

\bibitem[Deitke et~al.(2023)Deitke, Liu, Wallingford, Ngo, Michel, Kusupati, Fan, Laforte, Voleti, Gadre, VanderBilt, Kembhavi, Vondrick, Gkioxari, Ehsani, Schmidt, and Farhadi]{objaverseXL}
Matt Deitke, Ruoshi Liu, Matthew Wallingford, Huong Ngo, Oscar Michel, Aditya Kusupati, Alan Fan, Christian Laforte, Vikram Voleti, Samir~Yitzhak Gadre, Eli VanderBilt, Aniruddha Kembhavi, Carl Vondrick, Georgia Gkioxari, Kiana Ehsani, Ludwig Schmidt, and Ali Farhadi.
\newblock Objaverse-xl: A universe of 10m+ 3d objects.
\newblock \emph{arXiv preprint arXiv:2307.05663}, 2023.

\bibitem[Downs et~al.(2022)Downs, Francis, Koenig, Kinman, Hickman, Reymann, McHugh, and Vanhoucke]{downs2022google}
Laura Downs, Anthony Francis, Nate Koenig, Brandon Kinman, Ryan Hickman, Krista Reymann, Thomas~B McHugh, and Vincent Vanhoucke.
\newblock Google scanned objects: A high-quality dataset of 3d scanned household items.
\newblock In \emph{International Conference on Robotics and Automation (ICRA)}, pages 2553--2560. IEEE, 2022.

\bibitem[Gao et~al.(2023)Gao, Aigerman, Thibault, Kim, and Hanocka]{Gao_2023_TextDeformer}
William Gao, Noam Aigerman, Groueix Thibault, Vladimir Kim, and Rana Hanocka.
\newblock {TextDeformer: Geometry Manipulation using Text Guidance}.
\newblock In \emph{ACM Transactions on Graphics (SIGGRAPH)}, 2023.

\bibitem[Goel et~al.(2020)Goel, Kanazawa, and Malik]{ucmrGoel20}
Shubham Goel, Angjoo Kanazawa, and Jitendra Malik.
\newblock Shape and viewpoints without keypoints.
\newblock In \emph{ECCV}, 2020.

\bibitem[Heusel et~al.(2017)Heusel, Ramsauer, Unterthiner, Nessler, and Hochreiter]{FID}
Martin Heusel, Hubert Ramsauer, Thomas Unterthiner, Bernhard Nessler, and Sepp Hochreiter.
\newblock {GANs trained by a two time-scale update rule converge to a local nash equilibrium}.
\newblock \emph{Advances in neural information processing systems}, 30, 2017.

\bibitem[Ho et~al.(2020)Ho, Jain, and Abbeel]{DDPM}
Jonathan Ho, Ajay Jain, and Pieter Abbeel.
\newblock Denoising diffusion probabilistic models.
\newblock \emph{Advances in neural information processing systems}, 33:\penalty0 6840--6851, 2020.

\bibitem[Hong et~al.(2024)Hong, Zhang, Gu, Bi, Zhou, Liu, Liu, Sunkavalli, Bui, and Tan]{hong2023lrm}
Yicong Hong, Kai Zhang, Jiuxiang Gu, Sai Bi, Yang Zhou, Difan Liu, Feng Liu, Kalyan Sunkavalli, Trung Bui, and Hao Tan.
\newblock {LRM: Large reconstruction model for single image to 3D}.
\newblock In \emph{ICLR}, 2024.

\bibitem[Hu et~al.(2021)Hu, Wang, Xu, Liu, and Jia]{SMR}
Tao Hu, Liwei Wang, Xiaogang Xu, Shu Liu, and Jiaya Jia.
\newblock Self-supervised 3d mesh reconstruction from single images.
\newblock In \emph{CVPR}, pages 6002--6011, 2021.

\bibitem[Igarashi et~al.(2005)Igarashi, Moscovich, and Hughes]{igarashi2005ARAP}
Takeo Igarashi, Tomer Moscovich, and John~F Hughes.
\newblock As-rigid-as-possible shape manipulation.
\newblock \emph{ACM transactions on Graphics (TOG)}, 24\penalty0 (3):\penalty0 1134--1141, 2005.

\bibitem[Jakab et~al.(2021)Jakab, Tucker, Makadia, Wu, Snavely, and Kanazawa]{jakab2021keypointdeformer}
Tomas Jakab, Richard Tucker, Ameesh Makadia, Jiajun Wu, Noah Snavely, and Angjoo Kanazawa.
\newblock {Keypointdeformer: Unsupervised 3D keypoint discovery for shape control}.
\newblock In \emph{CVPR}, pages 12783--12792, 2021.

\bibitem[Kanazawa et~al.(2018)Kanazawa, Tulsiani, Efros, and Malik]{cmrKanazawa18}
Angjoo Kanazawa, Shubham Tulsiani, Alexei~A. Efros, and Jitendra Malik.
\newblock Learning category-specific mesh reconstruction from image collections.
\newblock In \emph{ECCV}, 2018.

\bibitem[Kerbl et~al.(2023)Kerbl, Kopanas, Leimk{\"u}hler, and Drettakis]{3DGS}
Bernhard Kerbl, Georgios Kopanas, Thomas Leimk{\"u}hler, and George Drettakis.
\newblock {3D Gaussian Splatting for Real-Time Radiance Field Rendering}.
\newblock \emph{ACM Transactions on Graphics}, 42\penalty0 (4), 2023.

\bibitem[Kim et~al.(2023)Kim, Angelina~Uy, Paschalidou, Jacobson, Guibas, and Sung]{kim2023optctrlpoints}
Kunho Kim, Mikaela Angelina~Uy, Despoina Paschalidou, Alec Jacobson, Leonidas~J Guibas, and Minhyuk Sung.
\newblock {OptCtrlPoints: Finding the Optimal Control Points for Biharmonic 3D Shape Deformation}.
\newblock In \emph{Computer Graphics Forum}, page e14963. Wiley Online Library, 2023.

\bibitem[Lan et~al.(2024)Lan, Hong, Yang, Zhou, Meng, Dai, Pan, and Loy]{lan2024ln3diff}
Yushi Lan, Fangzhou Hong, Shuai Yang, Shangchen Zhou, Xuyi Meng, Bo Dai, Xingang Pan, and Chen~Change Loy.
\newblock {LN3Diff: Scalable Latent Neural Fields Diffusion for Speedy 3D Generation}.
\newblock In \emph{ECCV}, 2024.

\bibitem[Li et~al.(2020)Li, Liu, Kim, De~Mello, Jampani, Yang, and Kautz]{umr2020}
Xueting Li, Sifei Liu, Kihwan Kim, Shalini De~Mello, Varun Jampani, Ming-Hsuan Yang, and Jan Kautz.
\newblock {Self-supervised Single-view 3D Reconstruction via Semantic Consistency}.
\newblock In \emph{ECCV}, 2020.

\bibitem[Lin et~al.(2023)Lin, Gao, Tang, Takikawa, Zeng, Huang, Kreis, Fidler, Liu, and Lin]{magic3d}
Chen-Hsuan Lin, Jun Gao, Luming Tang, Towaki Takikawa, Xiaohui Zeng, Xun Huang, Karsten Kreis, Sanja Fidler, Ming-Yu Liu, and Tsung-Yi Lin.
\newblock Magic3d: High-resolution text-to-3d content creation.
\newblock In \emph{CVPR}, 2023.

\bibitem[Lipman et~al.(2005)Lipman, Sorkine, Levin, and Cohen-Or]{lipman2005linear}
Yaron Lipman, Olga Sorkine, David Levin, and Daniel Cohen-Or.
\newblock Linear rotation-invariant coordinates for meshes.
\newblock \emph{ACM Transactions on Graphics (ToG)}, 24\penalty0 (3):\penalty0 479--487, 2005.

\bibitem[Liu et~al.(2023)Liu, Wu, Van~Hoorick, Tokmakov, Zakharov, and Vondrick]{liu2023zero1to3}
Ruoshi Liu, Rundi Wu, Basile Van~Hoorick, Pavel Tokmakov, Sergey Zakharov, and Carl Vondrick.
\newblock {Zero-1-to-3: Zero-shot one image to 3D object}.
\newblock In \emph{ICCV}, pages 9298--9309, 2023.

\bibitem[Liu et~al.(2024)Liu, Lin, Zeng, Long, Liu, Komura, and Wang]{liu2023syncdreamer}
Yuan Liu, Cheng Lin, Zijiao Zeng, Xiaoxiao Long, Lingjie Liu, Taku Komura, and Wenping Wang.
\newblock {SyncDreamer: Generating Multiview-consistent Images from a Single-view Image}.
\newblock In \emph{ICLR}, 2024.

\bibitem[Long et~al.(2024)Long, Guo, Lin, Liu, Dou, Liu, Ma, Zhang, Habermann, Theobalt, et~al.]{long2023wonder3d}
Xiaoxiao Long, Yuan-Chen Guo, Cheng Lin, Yuan Liu, Zhiyang Dou, Lingjie Liu, Yuexin Ma, Song-Hai Zhang, Marc Habermann, Christian Theobalt, et~al.
\newblock {Wonder3D: Single Image to 3D using Cross-Domain Diffusion}.
\newblock In \emph{CVPR}, 2024.

\bibitem[Mattes et~al.(2003)Mattes, Haynor, Vesselle, Lewellen, and Eubank]{deform_2D_1}
D. Mattes, D.R. Haynor, H. Vesselle, T.K. Lewellen, and W. Eubank.
\newblock {PET-CT image registration in the chest using free-form deformations}.
\newblock \emph{IEEE Transactions on Medical Imaging}, 22\penalty0 (1):\penalty0 120--128, 2003.

\bibitem[Mescheder et~al.(2019)Mescheder, Oechsle, Niemeyer, Nowozin, and Geiger]{onet}
Lars Mescheder, Michael Oechsle, Michael Niemeyer, Sebastian Nowozin, and Andreas Geiger.
\newblock Occupancy {Networks}: Learning {3D} reconstruction in function space.
\newblock In \emph{CVPR}, pages 4460--4470, 2019.

\bibitem[Mildenhall et~al.(2020)Mildenhall, Srinivasan, Tancik, Barron, Ramamoorthi, and Ng]{mildenhall2020nerf}
Ben Mildenhall, Pratul~P. Srinivasan, Matthew Tancik, Jonathan~T. Barron, Ravi Ramamoorthi, and Ren Ng.
\newblock {NeRF: Representing Scenes as Neural Radiance Fields for View Synthesis}.
\newblock In \emph{ECCV}, 2020.

\bibitem[Pang et~al.(2024)Pang, Jia, Shi, Tang, Zhang, Cheng, Zhou, Tay, and Yuan]{envision3d}
Yatian Pang, Tanghui Jia, Yujun Shi, Zhenyu Tang, Junwu Zhang, Xinhua Cheng, Xing Zhou, Francis E.~H. Tay, and Li Yuan.
\newblock Envision3d: One image to 3d with anchor views interpolation, 2024.

\bibitem[Park et~al.(2019)Park, Florence, Straub, Newcombe, and Lovegrove]{deepsdf}
Jeong~Joon Park, Peter Florence, Julian Straub, Richard Newcombe, and Steven Lovegrove.
\newblock {DeepSDF: Learning Continuous Signed Distance Functions for Shape Representation}.
\newblock In \emph{CVPR}, 2019.

\bibitem[Poole et~al.(2022)Poole, Jain, Barron, and Mildenhall]{poole2022dreamfusion}
Ben Poole, Ajay Jain, Jonathan~T. Barron, and Ben Mildenhall.
\newblock {DreamFusion: Text-to-3D using 2D Diffusion}.
\newblock \emph{arXiv preprint arXiv:2209.14988}, 2022.

\bibitem[Radford et~al.(2021)Radford, Kim, Hallacy, Ramesh, Goh, Agarwal, Sastry, Askell, Mishkin, Clark, et~al.]{clip}
Alec Radford, Jong~Wook Kim, Chris Hallacy, Aditya Ramesh, Gabriel Goh, Sandhini Agarwal, Girish Sastry, Amanda Askell, Pamela Mishkin, Jack Clark, et~al.
\newblock Learning transferable visual models from natural language supervision.
\newblock In \emph{International conference on machine learning}, pages 8748--8763. PMLR, 2021.

\bibitem[Shi et~al.(2023{\natexlab{a}})Shi, Chen, Zhang, Liu, Xu, Wei, Chen, Zeng, and Su]{shi2023zero123plus}
Ruoxi Shi, Hansheng Chen, Zhuoyang Zhang, Minghua Liu, Chao Xu, Xinyue Wei, Linghao Chen, Chong Zeng, and Hao Su.
\newblock Zero123++: a single image to consistent multi-view diffusion base model.
\newblock \emph{arXiv preprint arXiv:2310.15110}, 2023{\natexlab{a}}.

\bibitem[Shi et~al.(2023{\natexlab{b}})Shi, Wang, Ye, Mai, Li, and Yang]{shi2023MVDream}
Yichun Shi, Peng Wang, Jianglong Ye, Long Mai, Kejie Li, and Xiao Yang.
\newblock {MVDream: Multi-view Diffusion for 3D Generation}.
\newblock \emph{arXiv:2308.16512}, 2023{\natexlab{b}}.

\bibitem[Sorkine et~al.(2004)Sorkine, Cohen-Or, Lipman, Alexa, R{\"o}ssl, and Seidel]{sorkine2004laplacian}
Olga Sorkine, Daniel Cohen-Or, Yaron Lipman, Marc Alexa, Christian R{\"o}ssl, and H-P Seidel.
\newblock Laplacian surface editing.
\newblock In \emph{Proceedings of the 2004 Eurographics/ACM SIGGRAPH symposium on Geometry processing}, pages 175--184, 2004.

\bibitem[SUDO-AI-3D(2024)]{zero123_normal}
SUDO-AI-3D.
\newblock zero123plus/examples/normal\_gen.py at main.
\newblock \url{https://github.com/SUDO-AI-3D/zero123plus/blob/main/examples/normal_gen.py}, 2024.

\bibitem[Tang et~al.(2023)Tang, Wang, Zhang, Zhang, Yi, Ma, and Chen]{makeit3d}
Junshu Tang, Tengfei Wang, Bo Zhang, Ting Zhang, Ran Yi, Lizhuang Ma, and Dong Chen.
\newblock Make-it-3d: High-fidelity 3d creation from a single image with diffusion prior.
\newblock In \emph{ICCV}, pages 22819--22829, 2023.

\bibitem[Tang et~al.(2024{\natexlab{a}})Tang, Chen, Chen, Wang, Zeng, and Liu]{tang2024lgm}
Jiaxiang Tang, Zhaoxi Chen, Xiaokang Chen, Tengfei Wang, Gang Zeng, and Ziwei Liu.
\newblock {LGM: Large Multi-View Gaussian Model for High-Resolution 3D Content Creation}.
\newblock In \emph{ECCV}, pages 1--18. Springer, 2024{\natexlab{a}}.

\bibitem[Tang et~al.(2024{\natexlab{b}})Tang, Ren, Zhou, Liu, and Zeng]{dreamgaussian}
Jiaxiang Tang, Jiawei Ren, Hang Zhou, Ziwei Liu, and Gang Zeng.
\newblock Dreamgaussian: Generative gaussian splatting for efficient 3d content creation.
\newblock In \emph{ICLR}, 2024{\natexlab{b}}.

\bibitem[Tang et~al.(2024{\natexlab{c}})Tang, Zhang, Cheng, Yu, Feng, Pang, Lin, and Yuan]{tang2024cycle3d}
Zhenyu Tang, Junwu Zhang, Xinhua Cheng, Wangbo Yu, Chaoran Feng, Yatian Pang, Bin Lin, and Li Yuan.
\newblock {Cycle3D: High-quality and Consistent Image-to-3D Generation via Generation-Reconstruction Cycle}.
\newblock \emph{arXiv preprint arXiv:2407.19548}, 2024{\natexlab{c}}.

\bibitem[Tao et~al.(2022)Tao, Xiao, Qi, Cheng, and Ji]{digital_twin}
Fei Tao, Bin Xiao, Qinglin Qi, Jiangfeng Cheng, and Ping Ji.
\newblock Digital twin modeling.
\newblock \emph{Journal of Manufacturing Systems}, 64:\penalty0 372--389, 2022.

\bibitem[Tochilkin et~al.(2024)Tochilkin, Pankratz, Liu, Huang, , Letts, Li, Liang, Laforte, Jampani, and Cao]{TripoSR2024}
Dmitry Tochilkin, David Pankratz, Zexiang Liu, Zixuan Huang, , Adam Letts, Yangguang Li, Ding Liang, Christian Laforte, Varun Jampani, and Yan-Pei Cao.
\newblock {TripoSR: Fast 3D Object Reconstruction from a Single Image}.
\newblock \emph{arXiv preprint arXiv:2403.02151}, 2024.

\bibitem[Wang et~al.(2018)Wang, Zhang, Li, Fu, Liu, and Jiang]{wang2018pixel2mesh}
Nanyang Wang, Yinda Zhang, Zhuwen Li, Yanwei Fu, Wei Liu, and Yu-Gang Jiang.
\newblock {Pixel2Mesh: Generating 3D Mesh Models from Single RGB Images}.
\newblock In \emph{ECCV}, 2018.

\bibitem[Wang and Shi(2023)]{wang2023imagedream}
Peng Wang and Yichun Shi.
\newblock {ImageDream: Image-Prompt Multi-view Diffusion for 3D Generation}.
\newblock \emph{arXiv preprint arXiv:2312.02201}, 2023.

\bibitem[Wang et~al.(2021)Wang, Liu, Liu, Theobalt, Komura, and Wang]{wang2021neus}
Peng Wang, Lingjie Liu, Yuan Liu, Christian Theobalt, Taku Komura, and Wenping Wang.
\newblock {NeuS: Learning Neural Implicit Surfaces by Volume Rendering for Multi-view Reconstruction}.
\newblock \emph{Advances in Neural Information Processing Systems}, 34:\penalty0 27171--27183, 2021.

\bibitem[Wang et~al.(2022)Wang, Su, Zhang, Xing, Liu, Luan, and Shen]{metaverse}
Yuntao Wang, Zhou Su, Ning Zhang, Rui Xing, Dongxiao Liu, Tom~H Luan, and Xuemin Shen.
\newblock A survey on metaverse: Fundamentals, security, and privacy.
\newblock \emph{IEEE Communications Surveys \& Tutorials}, 25\penalty0 (1):\penalty0 319--352, 2022.

\bibitem[Wang et~al.(2004)Wang, Bovik, Sheikh, and Simoncelli]{SSIM}
Zhou Wang, Alan~C Bovik, Hamid~R Sheikh, and Eero~P Simoncelli.
\newblock Image quality assessment: from error visibility to structural similarity.
\newblock \emph{IEEE TIP}, 13\penalty0 (4):\penalty0 600--612, 2004.

\bibitem[Wang et~al.(2025)Wang, Wang, Chen, Xiang, Chen, Yu, Li, Su, and Zhu]{wang2025crm}
Zhengyi Wang, Yikai Wang, Yifei Chen, Chendong Xiang, Shuo Chen, Dajiang Yu, Chongxuan Li, Hang Su, and Jun Zhu.
\newblock {CRM: Single image to 3D textured mesh with convolutional reconstruction model}.
\newblock In \emph{ECCV}, pages 57--74. Springer, 2025.

\bibitem[Wu et~al.(2024)Wu, Liu, Cai, Yan, Wang, Hu, Duan, and Ma]{wu2024unique3d}
Kailu Wu, Fangfu Liu, Zhihan Cai, Runjie Yan, Hanyang Wang, Yating Hu, Yueqi Duan, and Kaisheng Ma.
\newblock {Unique3D: High-Quality and Efficient 3D Mesh Generation from a Single Image}.
\newblock In \emph{NeurIPS}, 2024.

\bibitem[Xu et~al.(2024)Xu, Cheng, Gao, Wang, Gao, and Shan]{xu2024instantmesh}
Jiale Xu, Weihao Cheng, Yiming Gao, Xintao Wang, Shenghua Gao, and Ying Shan.
\newblock {InstantMesh: Efficient 3D Mesh Generation from a Single Image with Sparse-view Large Reconstruction Models}.
\newblock \emph{arXiv preprint arXiv:2404.07191}, 2024.

\bibitem[Yang et~al.(2024)Yang, Chen, Pan, Yao, Chen, Ngo, and Mei]{yang2024hi3d}
Haibo Yang, Yang Chen, Yingwei Pan, Ting Yao, Zhineng Chen, Chong-Wah Ngo, and Tao Mei.
\newblock {Hi3D: Pursuing High-Resolution Image-to-3D Generation with Video Diffusion Models}.
\newblock In \emph{ACM MM}, 2024.

\bibitem[Yang et~al.(2023)Yang, Cheng, Duan, Ji, and Li]{yang2023consistnet}
Jiayu Yang, Ziang Cheng, Yunfei Duan, Pan Ji, and Hongdong Li.
\newblock {ConsistNet: Enforcing 3D Consistency for Multi-view Images Diffusion}.
\newblock \emph{arXiv preprint arXiv:2310.10343}, 2023.

\bibitem[Yoo et~al.(2024)Yoo, Kim, Kim, and Sung]{yoo2024apap}
Seungwoo Yoo, Kunho Kim, Vladimir~G. Kim, and Minhyuk Sung.
\newblock {As-Plausible-As-Possible: Plausibility-Aware Mesh Deformation Using 2D Diffusion Priors}.
\newblock In \emph{CVPR}, 2024.

\bibitem[Zhang et~al.(2024)Zhang, Tang, Pang, Cheng, Jin, Wei, Yu, Ning, and Yuan]{repaint123}
Junwu Zhang, Zhenyu Tang, Yatian Pang, Xinhua Cheng, Peng Jin, Yida Wei, Wangbo Yu, Munan Ning, and Li Yuan.
\newblock Repaint123: Fast and high-quality one image to 3d generation with progressive controllable 2d repainting.
\newblock In \emph{ECCV}, 2024.

\bibitem[Zhang et~al.(2018)Zhang, Isola, Efros, Shechtman, and Wang]{lpips}
Richard Zhang, Phillip Isola, Alexei~A Efros, Eli Shechtman, and Oliver Wang.
\newblock The unreasonable effectiveness of deep features as a perceptual metric.
\newblock In \emph{CVPR}, pages 586--595, 2018.

\end{thebibliography}
}

\appendix
\clearpage
\setcounter{page}{1}
\maketitlesupplementary

\section{Implementation details}
\subsection{Hyper parameters}
\label{sup_sec:hyper}
We use the same camera settings as InstantMesh~\cite{xu2024instantmesh}. The default camera FOV (Field of View) angle is 30°, and the default distance from the camera to the origin is 4. The multiview images are from six views, where the elevation angle is absolute and the azimuth angle is relative. The elevation angles are [20.0, -10.0, 20.0, -10.0, 20.0, -10.0] degrees, and the azimuth angles are [30.0, 90.0, 150.0, 210.0, 270.0, 330.0] degrees, respectively.

The resolution $G$ of the 2D deformation field grid is 20. We empirically set our loss weights $w_1$-$w_6$ for $\mathcal{L}_{\text{MSE}_1}$, $\mathcal{L}_{\text{mask}_1}$, $\mathcal{L}_{\text{smooth}_{2D}}$, $\mathcal{L}_{\text{MSE}_2}$, $\mathcal{L}_{\text{mask}_2}$, $\mathcal{L}_{\text{Lap}}$ to be 1.0, 1.0, 0.001, 1.0, 0.1, and 1e5, respectively. During the quantitative evaluation, the threshold for F-score is set to 0.2 following InstantMesh~\cite{xu2024instantmesh}.

\subsection{Geometry refinement}
As mentioned in the second-to-last sentence of \cref{sec:init}, we follow Unique3D~\cite{wu2024unique3d} to refine the geometry of the mesh. Specifically, we first generate multiview normal maps as reference. Then, we optimize mesh vertex coordinates to approximate reference normal maps. %

\para{Multiview normal map generation.} Any existing normal estimation method can be adopted here. Our backbone InstantMesh's associated multiview diffusion model is finetuned from Zero123++ v1.2~\cite{shi2023zero123plus}, which also provides a multiview normal generation model~\cite{zero123_normal}. This model creates multiview normal maps conditioned on both the input image and the generated multiview RGB images. Since InstantMesh uses white-background multiview RGB images, but the original Zero123++ uses gray-background ones for normal generation, so we first multiply the white-background images with a scale factor to make the whole image darker and the background gray, then feed the darker image to the normal generation pipeline. Although the RGB image is darker, we find that the generated normals are good. For the LGM~\cite{tang2024lgm} backbone in \cref{fig:other_backbone}, we use the normal estimation model~\cite{unique3d_normal} from Unique3D~\cite{wu2024unique3d} to generate multiview normal maps. Since both LGM and Unique3D use the same four views (front, back, left, right), Unique3D's trained normal diffusion model works well on LGM.  

\para{Mesh vertex optimization.} To refine the geometry, we optimize vertex vertices of the mesh to approximate the generated normal maps. Specifically, in each iteration, we render the mesh's normal maps, and calculate the following losses for backpropagation: MSE loss, mask loss, and expansion loss between rendered and reference normal maps, and a 3D Laplacian smooth loss. The MSE loss and the mask loss measure the average squared differences between the RGB and alpha channel values of corresponding pixels, respectively. The expansion loss is a regularization method proposed by Unique3D~\cite{wu2024unique3d}. It measures the mean squared difference between the original vertex positions and their positions after being moved along their normals. The 3D Laplacian smooth loss is introduced in \cref{equ:3D_lap}. The weights for MSE, mask, expansion and 3D Laplacian loss are 1.0, 1.0, 0.1, and 1e5, respectively. Please refer to Unique3D~\cite{wu2024unique3d} for more details.

\subsection{Camera pose estimation}
\label{sup_sec:cam_est}
As mentioned in ``Camera pose estimation" of \cref{sec:fidelity}, when the input image $I^\text{in}$ has a relatively high absolute elevation angle, we need to estimate $I^\text{in}$'s camera pose. Specifically, we first perform a coarse-to-fine grid search over all possible elevation angles, and then we further add a small optimization loop to further optimize the camera parameters. The details are presented  below:

\para{Coarse-to-fine search} Since our backbone InstantMesh uses relative azimuth and absolute elevation angles, we only need to find a suitable elevation. As for azimuth, simply setting it to zero would ensure the alignment between the generated mesh and the input image. For other camera parameters, we simply set them as the default setting as in \cref{sup_sec:hyper}. In the first stage (coarse), we search all elevation angles from -90° to 90° with the step of 3°. For each elevation angle, we use it to render our mesh, and calculate the LPIPS score between the rendered result and the input image. Among all angles, we adopt the angle $ele_1$ with the lowest LPIPS score, and feed it to the second stage. In the second (fine) stage, we search from $ele_1-3$ to $ele_1+3$ with the step of 1, and still adopt the angle with the lowest LPIPS score. 

\para{Optimization loop.} We optimize the camera position for 100 iterations to minimize the difference between the mesh's rendered result and the input image.

\subsection{Alignment with GT for evaluation.} As mentioned in the paragraph before ``qualitative results" in \cref{sec:compare_sota}, some methods do not align with GT by default when comparing rendered results and ground truth images for quantitative evaluation, so we conduct an alignment process for them. Specifically, in all our baselines, there are two methods that need this alignment: InstantMesh~\cite{xu2024instantmesh} and LGM~\cite{tang2024lgm}. 

\begin{itemize}
\item For InstantMesh, we directly adopt the method in \cref{sup_sec:cam_est}, since both InstantMesh and our Fancy123 use the same camera settings. 

    \item For LGM, we use a similar coarse-to-fine grid search strategy and choose the camera parameters with the lowest LPIPS score as in \cref{sup_sec:cam_est}, except for the following differences: 
    \begin{itemize}
        \item We search both elevation and azimuth angles for LGM, as both angles are absolute. 

\item In the first stage (coarse), we search elevation and azimuth angles from -90° to 90°, with the step size of 10°, resulting in $19 \times 19 = 361$ combinations. 

\item In the second stage (fine), we search angles around the best angles found in the first stage. Specifically, we search from the best elevation angle ± 9 degrees and the best azimuth angle ± 9 degrees, with a step size of 2.
    \end{itemize}

\end{itemize}

\subsection{Other mesh deformation methods}
As shown in \cref{fig:3D_deform_ablation}, we compare our Jacobian-field-based 3D mesh deformation with other two methods: vertex replacement and 3D deformation field. Here we provide details on these two methods.

\para{Vertex replacement:} Directly optimizing mesh vertex coordinates instead of Jacobian fields. The losses are the same as when using Jacobian fields. From \cref{fig:3D_deform_ablation}, we can see that this strategy fails to keep smoothness and plausibility. This is consistent with the observations in Fig. 3 of~\cite{Gao_2023_TextDeformer}, where the authors compare Jacobian-based and vertex-replacement-based mesh deformation and the former is significantly more globally coherent.

\para{3D deformation field:} Using 3D deformation fields as a 3D grid to control the deformation of each mesh vertex, similar to the 2D deformation field we adopt in our appearance enhancement module. We optimize the offset of each grid vertex, and the offset of each mesh vertex is the linear interpolation of nearby grid vertices. We can see in \cref{fig:3D_deform_ablation} that this strategy still cannot ensure global smooth and plausible deformation.

\section{More Quantitative results}
\subsection{elevation angle =elevation angle=0°}

As mentioned in \cref{sec:compare_sota} ``Datasets'', we provide results of using frontal views (elevation = azimuth = 0°) here in \cref{sup_tab:results_azi0}. Although our Fancy123 archives the best scores in 5 out of 7 metrics, we recommend focusing more on the qualitative results, see \cref{sup_sec:metric_discuss}.

\begin{table}[t]
  \centering
   \resizebox{1.0\linewidth}{!}{
    \begin{NiceTabular}{c|ccccc|cc}
    \toprule
          & \multicolumn{5}{c|}{Appearence Metrics} & \multicolumn{2}{c}{Geometry Metrics} \\
    \midrule
    Method & \multicolumn{1}{c}{FID ↓} & \multicolumn{1}{c}{LPIPS ↓} & \multicolumn{1}{c}{PSNR ↑} & \multicolumn{1}{c}{SSIM ↑} & \multicolumn{1}{c|}{CLIP-Sim. ↑} & \multicolumn{1}{c}{CD ↓} & \multicolumn{1}{c}{F-Score ↑} \\
    \midrule
    
    LGM~\cite{tang2024lgm}  & 81.29 & 0.4287 & 12.11 & 0.574 & 0.700 & 0.037 & 0.7605 \\

    TripoSR~\cite{TripoSR2024} & 71.65 & 0.3805 & 13.06 & 0.608 & 0.724 & 0.030 & 0.8047 \\

    CRM~\cite{wang2025crm}  & 82.18 & 0.4122 & 13.01 & 0.592 & 0.714 & 0.035 & 0.7768 \\
        
    InstantMesh~\cite{xu2024instantmesh} & \cellcolor[rgb]{ 1,  .788,  .78} 53.00 & 0.3671 & 13.25 & \cellcolor[rgb]{ 1,  .788,  .78} 0.617 & \cellcolor[rgb]{ 1,  .788,  .78} 0.780 & \cellcolor[rgb]{ 1,  .612,  .6} 0.023 & \cellcolor[rgb]{ 1,  .612,  .6} 0.8451 \\
    
    Unique3D~\cite{wu2024unique3d}  & 64.11 & 0.3821 & 12.79 & 0.601 & 0.754 & 0.033 & 0.7675 \\
    
    SF3D~\cite{sf3d2024}  & 58.15 & \cellcolor[rgb]{ 1,  .788,  .78} 0.3665 & \cellcolor[rgb]{ 1,  .788,  .78} 13.75 & 0.595 & 0.749 & 0.026 & 0.8279 \\

    \midrule
    Ours  & \cellcolor[rgb]{ 1,  .612,  .6} 46.05 & \cellcolor[rgb]{ 1,  .612,  .6} 0.3521 & \cellcolor[rgb]{ 1,  .612,  .6} 13.79 & \cellcolor[rgb]{ 1,  .612,  .6} 0.633 & \cellcolor[rgb]{ 1,  .612,  .6} 0.803 & \cellcolor[rgb]{ 1,  .788,  .78} 0.024 & \cellcolor[rgb]{ 1,  .788,  .78} 0.8384 \\
    \bottomrule
    \end{NiceTabular}%
    }
   
    \caption{Quantitative comparisons of our method against baseline methods (ranked by initial paper release time) for the single-image-to-3D-mesh task.  }
     \vspace{-1em}
  \label{sup_tab:results_azi0}%
\end{table}%

\subsection{Inadequacies of existing quantitative metrics}
\label{sup_sec:metric_discuss}
As mentioned in \cref{sec:compare_sota} ``Quantitative results", the quantitative metrics often do not align with human perception. 

\begin{figure*}[t]
    \centering
    \includegraphics[width=1.0\linewidth]{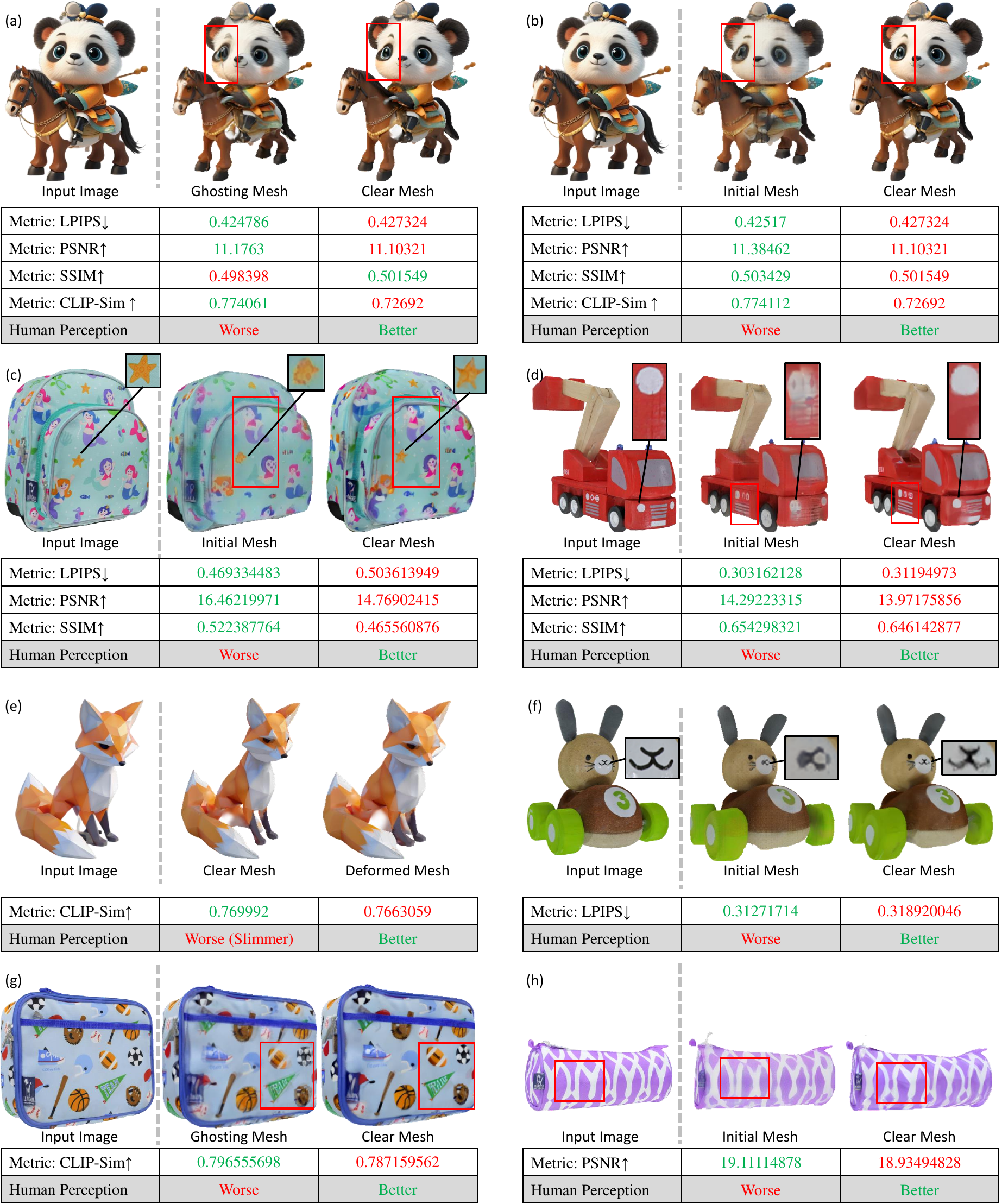}
    \caption{Illustration of the first limitation of existing quantitative metrics: mismatch with human perception. From (a) to (h), each sample shows the input image and rendering results of two different generated meshes. The mesh on the right is better by human perception, without blurring or ghosting. The tables below list common metrics calculated by comparing the rendered and input images, with the last row showing the comparison results of human perception. The metrics do not align with human perception.%
    }
    \label{fig:bad_metric}
\end{figure*}

\para{Misalignment between metrics and human perception.} \cref{fig:bad_metric} shows more examples. From (a) to (h), each sample shows the input image and rendering results of two different generated meshes.

\begin{itemize}
    \item Initial Mesh: mesh generated by InstantMesh.
    \item Ghosting Mesh: unprojecting multiview images without 2D deformation to initial mesh.
    \item Clear Mesh: unproejcting multiview images after 2D deformation to the initial mesh.
    \item Deformed Mesh: deforming the clear mesh to match the input image.
\end{itemize}

Under each method's generated mesh's rendering result, we list the metrics comparing this very image and the corresponding input image (GT), together with human perception results. We can see that, for each sample, the mesh on the right is better by human perception, without blurring or ghosting, but the metrics disagree. Specifically: (1) the metrics only measure ``similarity to GT'' while ignoring significant artifacts, e.g. the thin black line under the eye in the red box of (a). (2) Even only considering ``similarity to GT'', the metrics can differ from human perception, see (e) where ``Clear Mesh" is too slim but its score is higher.

\begin{figure*}[t] %
    \centering
    \includegraphics[width=1.0\linewidth]{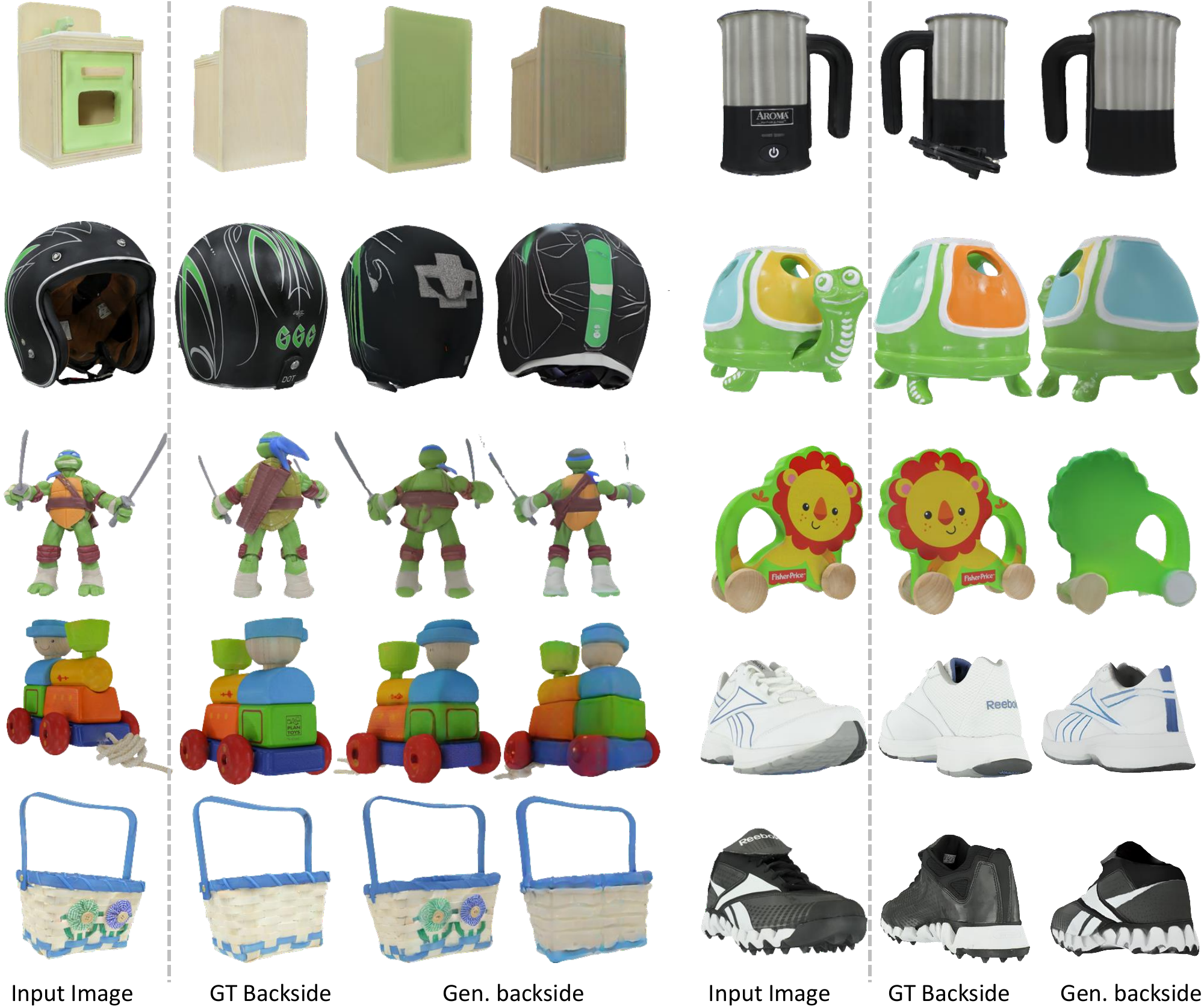}
    \caption{Illustration of the second limitation of existing quantitative metrics: the contradiction between the ambiguous task and the unique GT. For each sample, we display the input image, the ground-truth mesh's backside (GT Backside), and the backside of generated meshes (Gen. backside) from different methods including ours and baseline methods. We can see that, while there are different possibilities for the backside, the GT is unique. Therefore, existing metrics are improper, since they only compare results with the unique GT.
    }
    \label{fig:unique_GT}
\end{figure*}

\para{Contradiction between the ambiguous task and the unique GT.} For the single-image-to-3D task, the input only contains one image from a single view, so the generated mesh's novel views have various possibilities. However, existing metrics use one unique GT mesh for comparison, which is thus improper. \cref{fig:unique_GT} illustrates some examples of various possible backsides from a single-view image. Under such circumstances, though different backsides are plausible and would be equally treated by human perception, their metrics would differ significantly, as the metrics only compare with a unique GT.

\clearpage

\clearpage

\clearpage

\begin{figure*}[t] 
    \centering
    \includegraphics[width=1.0\linewidth]{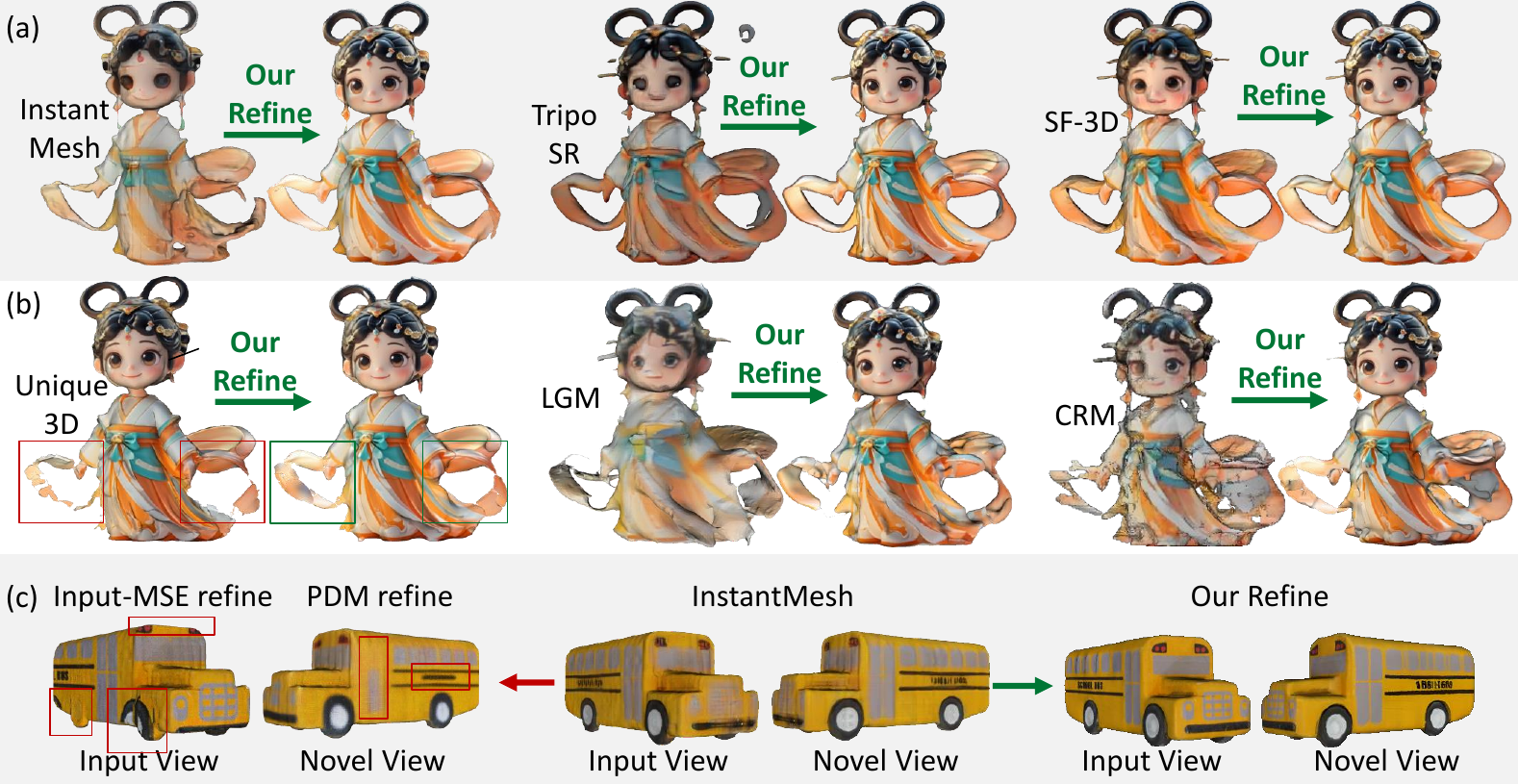}
    \caption{(a)-(b) Applying our Fancy123's enhancement modules to different initialization methods. (b) Fancy123 improves the results even for relatively low-quality initial meshes. (c) Comparison of Fancy123's refinement performance against Other refinement techniques. 
    }
    \label{fig:rebuttal}
\end{figure*}

\section{Initialization method and enhancement module replacement experiment}
To validate the universal applicability of the enhancement modules in Fancy123, we replace the mesh initialization method InstantMesh with other baseline methods, and apply Fancy123 enhancement modules as follows:

(1) TripoSR/SF3D + our enhancement: Since neither TripoSR nor SF3D generates multiview images as intermediate products, Fancy123's appearance enhancement module can not be applied. Thus, only geometry and fidelity enhancements are applied.

(2) Unique3D + our enhancement: Since Unique3D already incorporates geometry enhancement identical to Fancy123, only appearance and fidelity enhancements are applied.

(3) LGM/CRM/InstantMesh + our enhancement: We apply out complete enhancement pipeline (geometry, appearance, and fidelity) to meshes generated by LGM, CRM, and InstantMesh. For LGM and CRM, we generate multiview normal maps by the normal estimation model provided by Unique3D. All these three methods use the same four viewpoints (front, back, left, right), ensuring good model compatibility.
As shown in~\cref{fig:rebuttal} (a) and (b), our enhancement modules improved the generation quality for all initialization methods. Even for low-quality initial meshes in ~\cref{fig:rebuttal} (b) with geometric flaws (e.g., marked by red boxes ), Fancy123 enhances coherence, smoothness, and overall plausibility. 

Additionally, we compare Fancy123 with two typical enhancement techniques for single-image-to-3D:

(1) Input-MSE Enhancement: This intuitive approach optimizes mesh parameters by minimizing the mean squared error (MSE) loss between input images and mesh renderings. It is widely adopted in image-conditioned 3D generation (e.g., Make-It-3D~\cite{makeit3d}, DreamGaussian~\cite{dreamgaussian}), and it only enhances regions visible from the input viewpoint. However, it produces significant artifacts when geometric misalignment exists between the mesh and input image, as shown in ~\cref{fig:rebuttal} (c) ``Input-MSE refine'. In contrast, our Fancy123 achieves precise alignment through 3D deformation, avoiding such artifacts.

(2) SDS/PDM (Perturb-Denoise-MSE) Enhancement: SDS~\cite{poole2022dreamfusion} is a foundational method for "2D diffusion for 3D generation". It's also commonly used in refinement stages~\cite{magic3d}. It works by manually adding noise to mesh renderings and optimizing mesh parameters via SDS loss. DreamGaussian~\cite{dreamgaussian} notes that SDS-based UV color optimization introduces unwanted artifacts and proposes to replace SDS loss with MSE after perturbation, which we term ``PDM''. While both SDS and PDM enhance the whole meshes across all viewpoints, they suffer from texture blurring due to inconsistent 3D guidance signals inherent in diffusion denoising, as mentioned by DreamGaussian~\cite{dreamgaussian} authors. As shown in ~\cref{fig:rebuttal} (c) ``PDM refine'', PDM may yield an even blurrier texture than the initial mesh.

Quantitative results in~\cref{tab:rebuttal} show that our Fancy123 enhances all baselines, while Input-MSE and PDM degrade most metrics (green/red text indicating performance improvement/decline). We assume the key reasons are as mentioned above: (1) Input-MSE fails to handle geometric misalignment between 3D meshes and input images. (2) PDM reduces texture clarity due to inconsistent denoising guidance.

\begin{table}[t]
  \centering
   \resizebox{1.0\linewidth}{!}{
    \begin{tabular}{lccccc}
    \toprule
          & \multicolumn{1}{c}{FID ↓} & \multicolumn{1}{c}{LPIPS ↓} & \multicolumn{1}{c}{PSNR ↑} & \multicolumn{1}{c}{SSIM ↑} & \multicolumn{1}{c}{CLIP-Sim. ↑} \\
    \midrule
    LGM   & 86.24 & 0.395 & 13.254 & 0.612 & 0.710 \\
    \textbf{+our refine} & \textcolor[rgb]{ 0,  .69,  .314}{\textbf{58.92}} & \textcolor[rgb]{ 0,  .69,  .314}{\textbf{0.377}} & \textcolor[rgb]{ 0,  .69,  .314}{\textbf{13.266}} & \textcolor[rgb]{ .753,  0,  0}{\textbf{0.610}} & \textcolor[rgb]{ 0,  .69,  .314}{\textbf{0.748}} \\
        +input-MSE refine & \textcolor[rgb]{ 0,  .69,  .314}{69.44} & \textcolor[rgb]{ .753,  0,  0}{0.407} & \textcolor[rgb]{ .753,  0,  0}{12.745} & \textcolor[rgb]{ .753,  0,  0}{0.575} & \textcolor[rgb]{ .753,  0,  0}{0.696} \\
        +PDM refine & \textcolor[rgb]{ .753,  0,  0}{102.80} & \textcolor[rgb]{ .753,  0,  0}{0.397} & \textcolor[rgb]{ .753,  0,  0}{13.163} & \textcolor[rgb]{ .753,  0,  0}{0.599} & \textcolor[rgb]{ .753,  0,  0}{0.669} \\
    \midrule
    CRM   & 87.74 & 0.388 & 13.664 & 0.620 & 0.729 \\
    \textbf{    +our refine} & \textcolor[rgb]{ 0,  .69,  .314}{\textbf{57.65}} & \textcolor[rgb]{ 0,  .69,  .314}{\textbf{0.368}} & \textcolor[rgb]{ 0,  .69,  .314}{\textbf{13.756}} & \textcolor[rgb]{ 0,  .69,  .314}{\textbf{0.626}} & \textcolor[rgb]{ 0,  .69,  .314}{\textbf{0.784}} \\
        +input-MSE refine & \textcolor[rgb]{ 0,  .69,  .314}{75.31} & \textcolor[rgb]{ .753,  0,  0}{0.389} & \textcolor[rgb]{ 0,  .69,  .314}{13.669} & \textcolor[rgb]{ .753,  0,  0}{0.590} & \textcolor[rgb]{ .753,  0,  0}{0.725} \\
        +PDM refine & \textcolor[rgb]{ .753,  0,  0}{102.18} & \textcolor[rgb]{ .753,  0,  0}{0.391} & \textcolor[rgb]{ .753,  0,  0}{13.570} & \textcolor[rgb]{ .753,  0,  0}{0.594} & \textcolor[rgb]{ .753,  0,  0}{0.686} \\
    \midrule
    TripoSR & 68.52 & 0.367 & 13.876 & 0.629 & 0.739 \\
    \textbf{    +our refine} & \textcolor[rgb]{ 0,  .69,  .314}{\textbf{51.76}} & \textcolor[rgb]{ 0,  .69,  .314}{\textbf{0.347}} & \textcolor[rgb]{ 0,  .69,  .314}{\textbf{14.144}} & \textcolor[rgb]{ 0,  .69,  .314}{\textbf{0.635}} & \textcolor[rgb]{ 0,  .69,  .314}{\textbf{0.780}} \\
        +input-MSE refine & \textcolor[rgb]{ 0,  .69,  .314}{66.96} & \textcolor[rgb]{ .753,  0,  0}{0.369} & \textcolor[rgb]{ 0,  .69,  .314}{13.880} & \textcolor[rgb]{ .753,  0,  0}{0.619} & \textcolor[rgb]{ .753,  0,  0}{0.726} \\
        +PDM refine & \textcolor[rgb]{ .753,  0,  0}{89.82} & \textcolor[rgb]{ .753,  0,  0}{0.376} & \textcolor[rgb]{ .753,  0,  0}{13.934} & \textcolor[rgb]{ .753,  0,  0}{0.591} & \textcolor[rgb]{ .753,  0,  0}{0.703} \\
    \midrule
    InstantMesh & 46.16 & 0.349 & 13.735 & 0.634 & 0.800 \\
    \textbf{    +our refine} & \textcolor[rgb]{ 0,  .69,  .314}{\textbf{37.99}} & \textcolor[rgb]{ 0,  .69,  .314}{\textbf{0.330}} & \textcolor[rgb]{ 0,  .69,  .314}{\textbf{14.365}} & \textcolor[rgb]{ 0,  .69,  .314}{\textbf{0.651}} & \textcolor[rgb]{ 0,  .69,  .314}{\textbf{0.835}} \\
        +input-MSE refine & \textcolor[rgb]{ 0,  .69,  .314}{40.82} & \textcolor[rgb]{ .753,  0,  0}{0.354} & \textcolor[rgb]{ 0,  .69,  .314}{13.848} & \textcolor[rgb]{ .753,  0,  0}{0.619} & \textcolor[rgb]{ 0,  .69,  .314}{0.803} \\
        +PDM refine & \textcolor[rgb]{ .753,  0,  0}{72.94} & \textcolor[rgb]{ .753,  0,  0}{0.377} & \textcolor[rgb]{ 0,  .69,  .314}{13.837} & \textcolor[rgb]{ .753,  0,  0}{0.596} & \textcolor[rgb]{ .753,  0,  0}{0.734} \\
    \midrule
    Unique3D & 58.51 & 0.389 & 12.741 & 0.592 & 0.764 \\
    \textbf{    +our refine} & \textcolor[rgb]{ 0,  .69,  .314}{\textbf{53.69}} & \textcolor[rgb]{ 0,  .69,  .314}{\textbf{0.379}} & \textcolor[rgb]{ 0,  .69,  .314}{\textbf{13.094}} & \textcolor[rgb]{ 0,  .69,  .314}{\textbf{0.597}} & \textcolor[rgb]{ 0,  .69,  .314}{\textbf{0.774}} \\
        +input-MSE refine & \textcolor[rgb]{ 0,  .69,  .314}{57.66} & \textcolor[rgb]{ .753,  0,  0}{0.396} & \textcolor[rgb]{ 0,  .69,  .314}{12.867} & \textcolor[rgb]{ .753,  0,  0}{0.583} & \textcolor[rgb]{ .753,  0,  0}{0.742} \\
        +PDM refine & \textcolor[rgb]{ .753,  0,  0}{97.35} & \textcolor[rgb]{ .753,  0,  0}{0.414} & \textcolor[rgb]{ 0,  .69,  .314}{12.918} & \textcolor[rgb]{ .753,  0,  0}{0.558} & \textcolor[rgb]{ .753,  0,  0}{0.662} \\
    \midrule
    SF3D  & 49.89 & 0.343 & 14.320 & 0.617 & 0.776 \\
    \textbf{    +our refine} & \textcolor[rgb]{ .753,  0,  0}{\textbf{50.66}} & \textcolor[rgb]{ 0,  .69,  .314}{\textbf{0.332}} & \textcolor[rgb]{ 0,  .69,  .314}{\textbf{15.091}} & \textcolor[rgb]{ 0,  .69,  .314}{\textbf{0.648}} & \textcolor[rgb]{ 0,  .69,  .314}{\textbf{0.790}} \\
        +input-MSE refine & \textcolor[rgb]{ 0,  .69,  .314}{49.71} & \textcolor[rgb]{ .753,  0,  0}{0.348} & \textcolor[rgb]{ 0,  .69,  .314}{14.685} & \textcolor[rgb]{ .753,  0,  0}{0.615} & \textcolor[rgb]{ .753,  0,  0}{0.769} \\
        +PDM refine & \textcolor[rgb]{ .753,  0,  0}{79.77} & \textcolor[rgb]{ .753,  0,  0}{0.359} & \textcolor[rgb]{ .753,  0,  0}{14.220} & \textcolor[rgb]{ 0,  .69,  .314}{0.621} & \textcolor[rgb]{ .753,  0,  0}{0.728} \\
    \bottomrule
    \end{tabular}%
    }
    \caption{Results of different initialization methods without or with different refinement methods. Green and red indicate better and worse performance after refinement, respectively.}
  \label{tab:rebuttal}%
\end{table}%

\section{Can the order of enhancement modules be swapped?}
No. (1) The initial mesh without appearance enhancement (AE) may be too blurry for fidelity enhancement (FE), since 3D deformation in fidelity enhancement is guided by RGB loss. (2) AE unprojects the deformed multiview images onto the mesh, and FE unprojects the input image. If we first apply FE and then AE, the result of FE will be overwritten by AE. So we recommend following the order in our paper to apply AM then FM, so as to first address multiview inconsistency (across all views) by AE then improve the input view quality by FE.

\section{Failur case}
\begin{figure}[t] 
    \centering
    \includegraphics[width=1.0\linewidth]{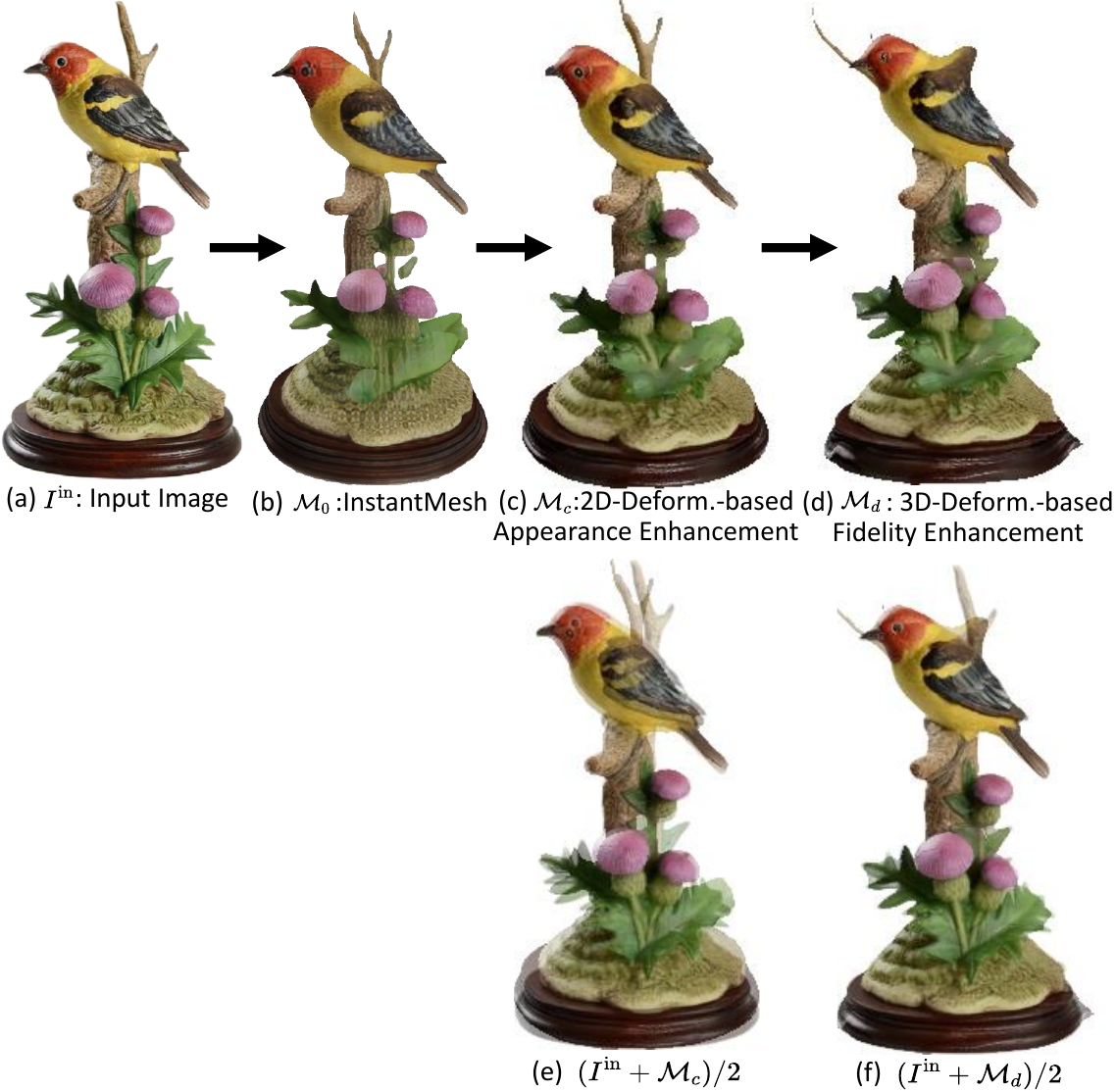}
    \caption{A failure case of our Fancy123 when different semantic parts share similar colors. When deforming mesh (c) $\mathcal{M}_c$ into (d) $\mathcal{M}_d$, the bird's brown back is mistakenly regarded as part of the brown branch, and therefore is incorrectly raised. %
    }
    \label{fig:failure_case}
\end{figure}

As mentioned in our manuscript, our 3D deformation module can sometimes lead to artifacts when different semantic parts of an object share similar colors. \cref{fig:failure_case} shows an example. When deforming mesh (c) $\mathcal{M}_c$ into (d) $\mathcal{M}_d$ to match the input image (a) $I^\text{in}$, the bird's brown back in (d) is mistakenly regarded as part of a branch and is therefore incorrectly raised. Our analysis is as follows:

\begin{itemize}
    \item From \cref{fig:failure_case} (e), we can see that the branch in the rendered image of $\mathcal{M}_c$. almost completely misaligns with the branch in $I^\text{in}$: the branch in $I^\text{in}$ is on the right of the branch in $\mathcal{M}_c$. 
    \item In \cref{fig:failure_case} (e) and (c), immediately below the correct branch position (on the right side) is the bird's brown back, which has a similar color to the branch. 
    \item Therefore, the 3D deformation optimization loop chooses to raise the bird's brown back to better match the input image, rather than moving the almost completely misaligned branch. 
    \item As for the branch, the optimization loop adopts a locally optimal approach: minimizing or even eliminating the branch, so as to reduce its pixel count and thus lower the loss function value.
To address such issues, we plan to further introduce semantic guidance in our 3D deformation module in the future. For now, we recommend skipping the 3D-deformation-based fidelity enhancement module, and only applying the 2D-deformation-based appearance enhancement module. 

\end{itemize}
\clearpage

\section{More qualitative results}

\begin{figure*}[t]
    \centering
    \includegraphics[width=1.0\linewidth]{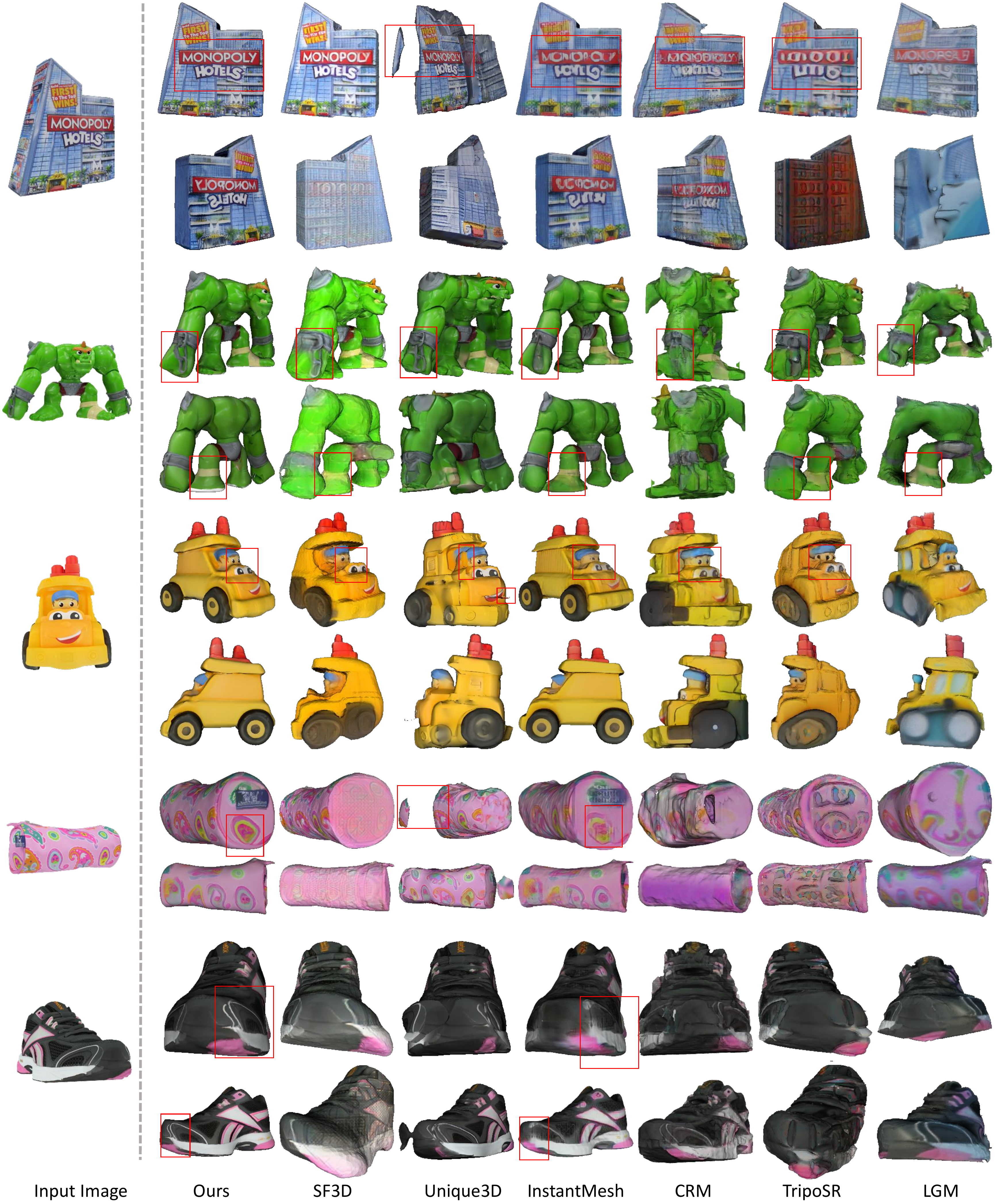}
    \caption{More visual comparisons. Our method exhibits high clarity and plausibility compared to baseline methods. Better zoom in.
    }
    \label{fig:more_results_1}
\end{figure*}

\begin{figure*}[t]
    \centering
    \includegraphics[width=1.0\linewidth]{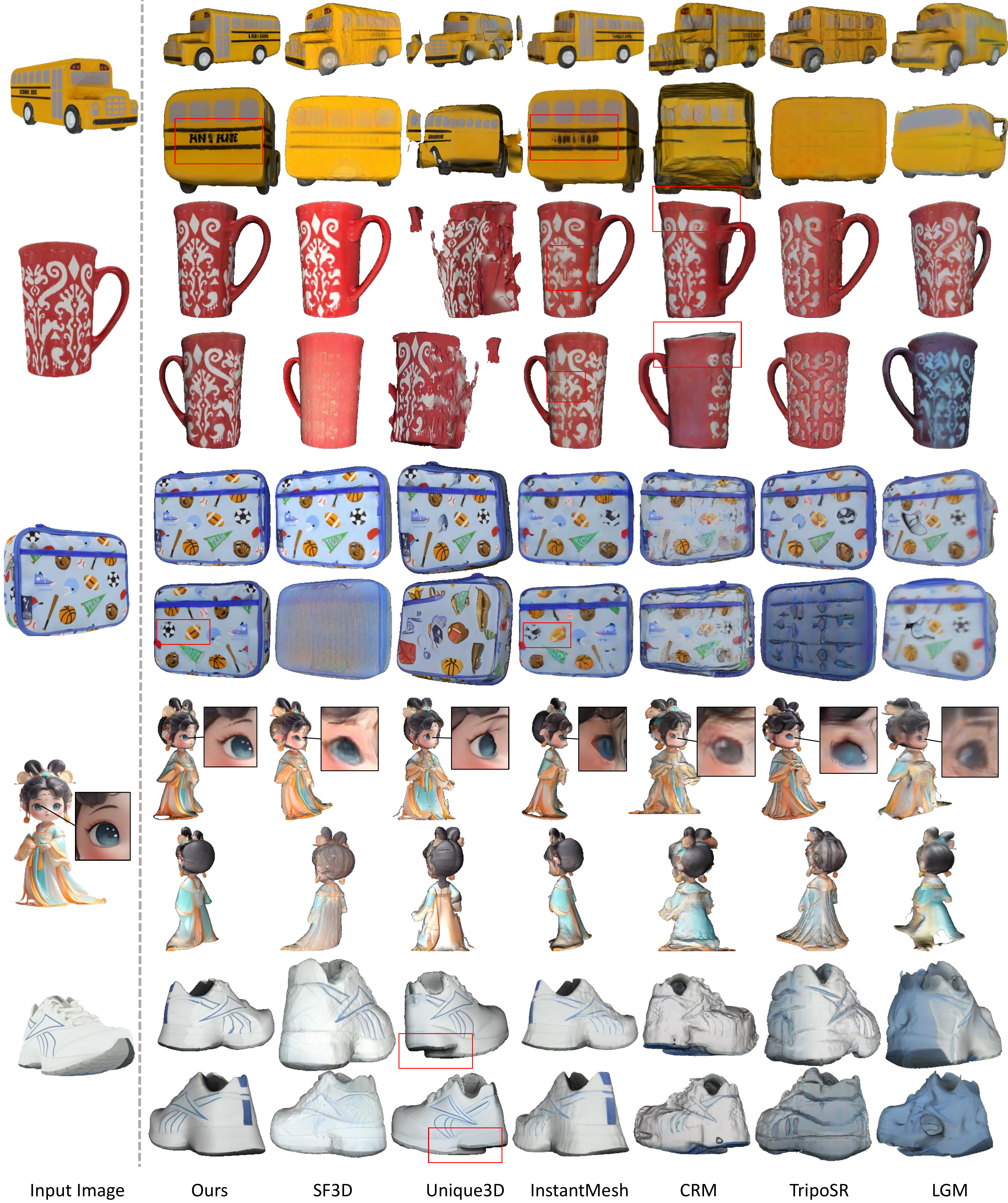}
    \caption{More visual comparisons. Our method exhibits high clarity and plausibility compared to baseline methods. Better zoom in.
    }
    \label{fig:more_results_2}
\end{figure*}

\subsection{Comparison with SoTA}
We provide more visual comparisons between our Fancy123 and baseline methods in \cref{fig:more_results_1,fig:more_results_2}. InstantMesh, CRM, TripoSR, and LGM often produce blurry-looking meshes. SF3D and TripoSR exhibit lower plausibility in novel views. Unique3D often yields significant artifacts as shown in the red boxes. Unlike them, our Fancy123 archives high clarity and plausibility for various challenging input images.

\subsection{Ablation study}
\begin{figure*}[t] %
    \centering
    \includegraphics[width=0.9\linewidth]{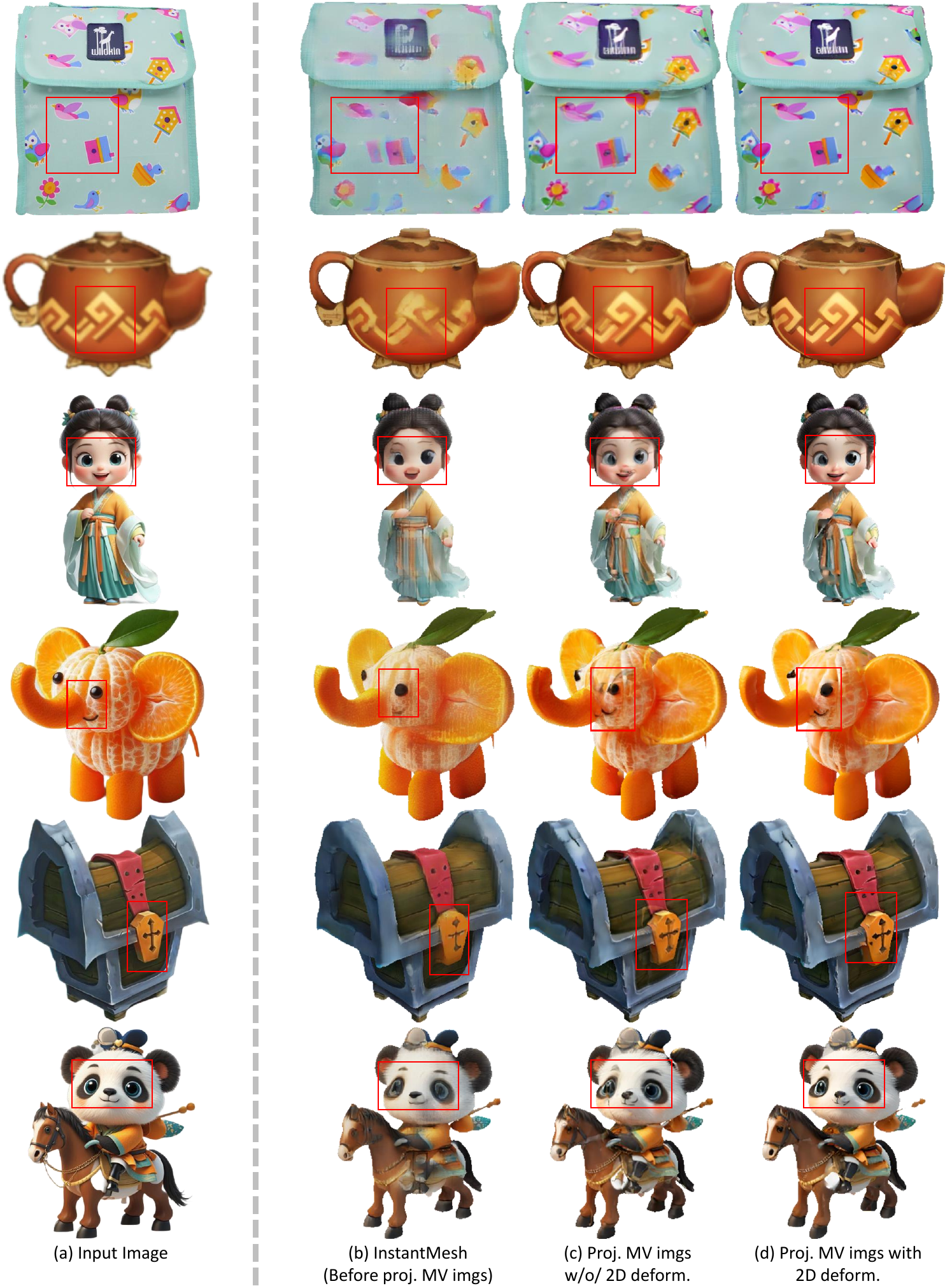}
    \caption{Ablation experiments on the 2D-deformation-based appearance enhancement module: unprojecting the deformed multiview images to the mesh (d) archives clear-looking mesh without blurring (b) or ghosting (c).
    }
    \label{fig:2D_ablation_1}
\end{figure*}

\begin{figure*}[t] 
    \centering
    \includegraphics[width=0.9\linewidth]{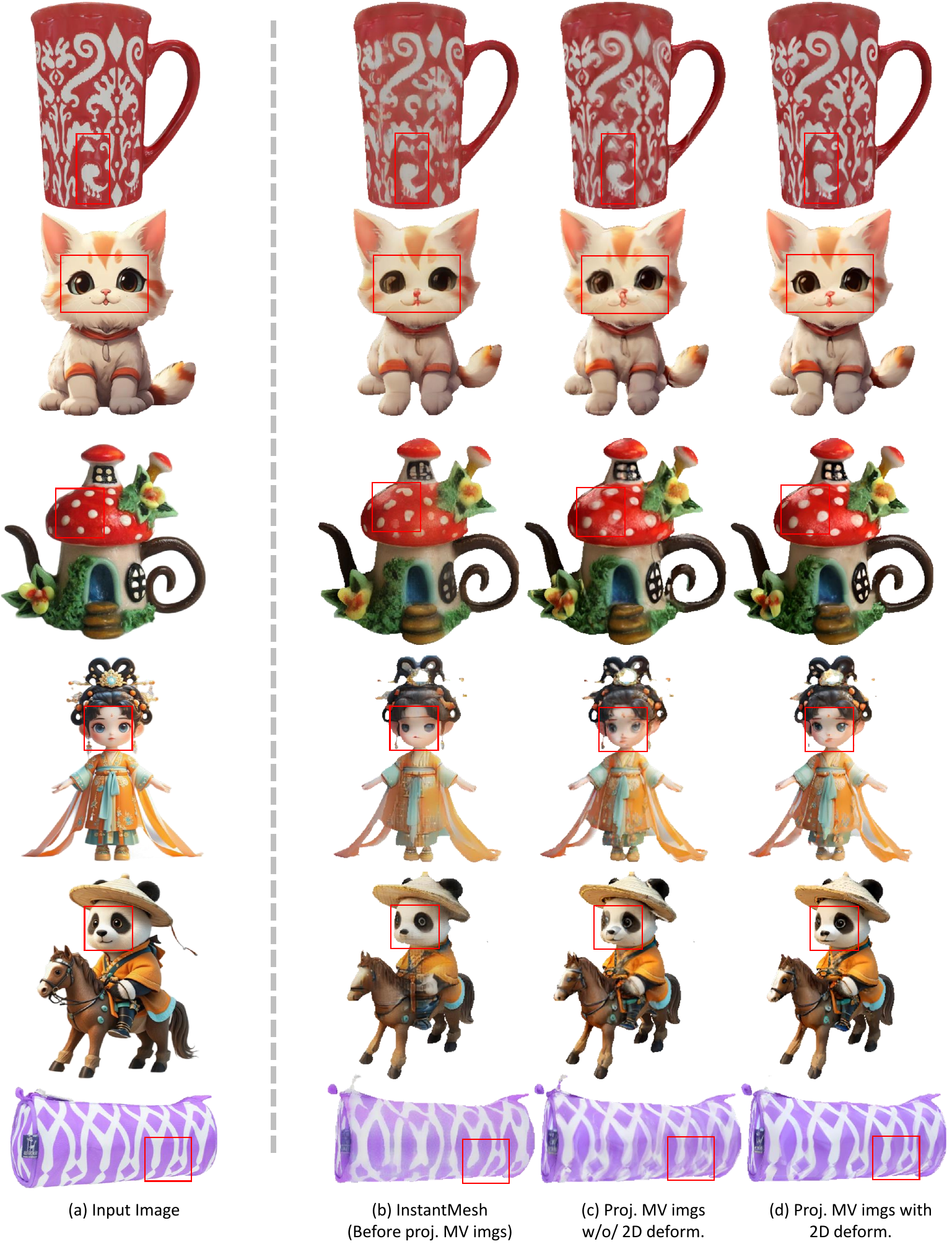}
    \caption{Ablation experiments on the 2D-deformation-based appearance enhancement module: unprojecting the deformed multiview images to the mesh (d) archives clear-looking mesh without blurring (b) or ghosting (c).
    }
    \label{fig:2D_ablation_2}
\end{figure*}

\para{2D deformation.} 
\cref{fig:2D_ablation_1,fig:2D_ablation_2} present more visual results on the effect of our 2D-deformation-based appearance enhancement module. 
\begin{itemize}
    \item (b): The original mesh generated by InstantMesh looks blurry.
    \item (c): Directly unprojecting the multiview images to the mesh without 2D deformation leads to ghosting.
    \item (d): After our appearance enhancement module: unprojecting the deformed multiview images to the mesh improves appearance quality, especially clarity.
\end{itemize}

\begin{figure*}[t] 
    \centering
    \includegraphics[width=1.0\linewidth]{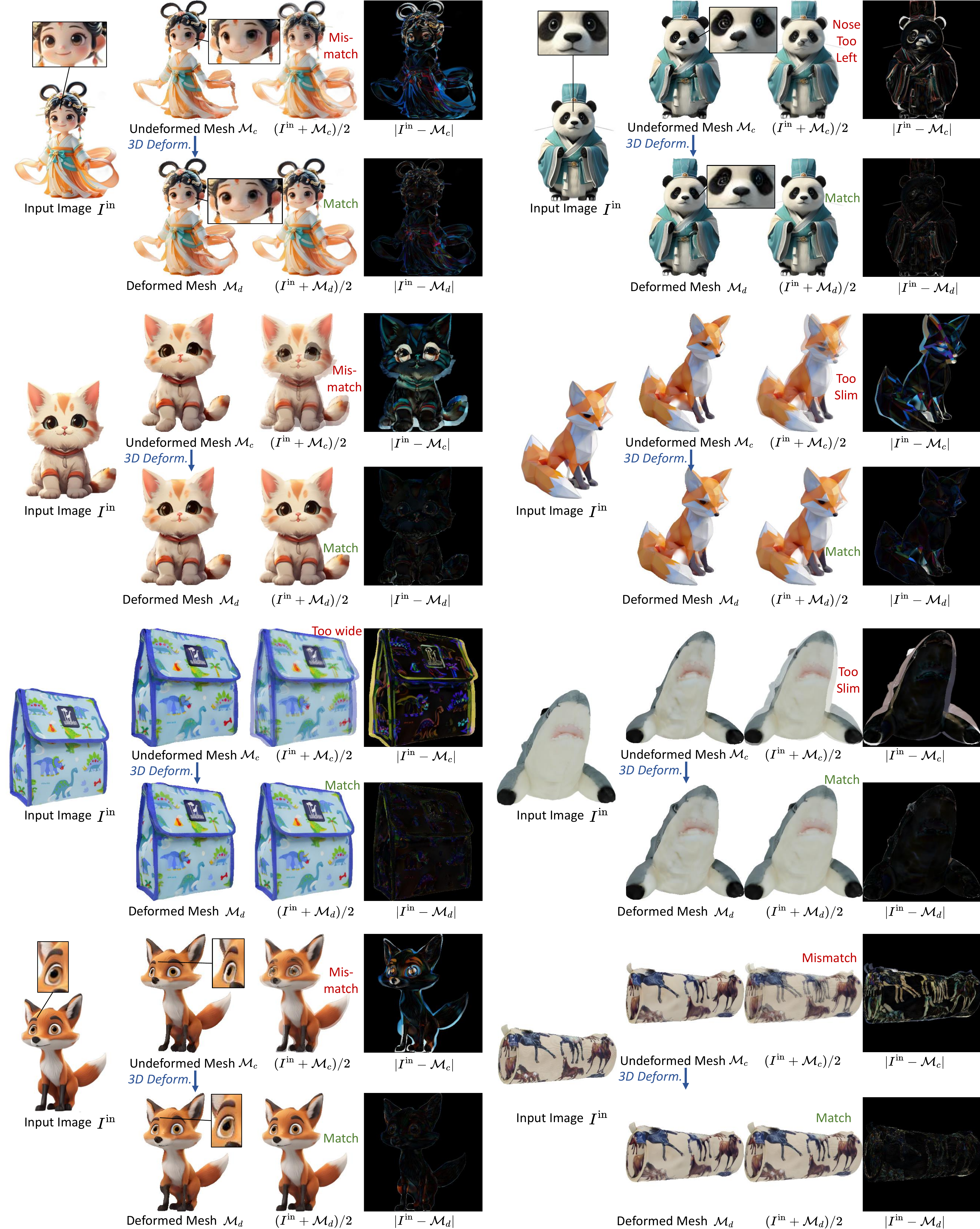}
    \caption{Ablation experiments on 3D mesh deformation. For simplicity, $\mathcal{M}_c$ and $\mathcal{M}_d$ denote the rendered images of meshes before and after 3D deformation, respectively, omitting the rendering symbol $\mathcal{R}$. By comparing $\mathcal{M}_c$ and $\mathcal{M}_d$ with the input image $I^\text{in}$ through addition or subtraction of them, we can see that  $\mathcal{M}_d$ matches $I^\text{in}$ better than $\mathcal{M}_c$, thus paving the path for the unprojection of $I^\text{in}$ to $\mathcal{M}_d$. 
    }
    \label{fig:3D_ablation_supp}
\end{figure*}

\para{3D deformation.} \cref{fig:3D_ablation_supp} illustrates more examples on the effect of our 3D deformation operation. For simplicity, we directly use the mesh symbol $\mathcal{M}$ to denote the rendered images of a mesh, omitting the rendering symbol $\mathcal{R}$.
\begin{itemize}
    \item $\mathcal{M}_c$: the rendered image of the mesh before 3D deformation.
    \item $\mathcal{M}_d$: the rendered image of the mesh after 3D deformation.
\end{itemize}

To measure the matching degree between a mesh $\mathcal{M}$ and the input image $I^\text{in}$, we perform two different forms of visualization: image overlay and subtraction:

\begin{itemize}
    \item $(I^\text{in} + \mathcal{M})/2$: Overlay two images. The ghosting regions indicate mismatches.
    \item $|I^\text{in} - \mathcal{M}|$: Subtract two images. The non-black regions indicate mismatches, with brighter areas indicating greater mismatches.
\end{itemize}
From \cref{fig:3D_ablation_supp}, we can see that, the undeformed mesh $\mathcal{M}_c$ often fails to match the input image $I^\text{in}$, which prevents us from unprojecting $I^\text{in}$ to $\mathcal{M}_c$. After 3D deformation, this mismatch issue is greatly alleviated. In other words, the mesh's fidelity to the input image is greatly improved.

\end{document}